\ifdefined\isarxivversion
\documentclass[11pt]{article}
\else

\documentclass[11pt,a4paper]{article}
\usepackage[hyperref]{emnlp2020}
\usepackage{times}
\usepackage{latexsym}

\aclfinalcopy
\fi

\usepackage{graphicx}
\usepackage{natbib}
\usepackage{graphicx}
\usepackage{comment}
\usepackage{amsmath}
\usepackage{amsthm}
\usepackage{amssymb}
\usepackage{array}
\usepackage{float}
\usepackage{subfig}
\usepackage{algorithm}
\usepackage{algpseudocode}
\usepackage{tikz}
\usepackage{comment}
\usepackage{hyperref}
\usepackage{multirow}
\usepackage[makeroom]{cancel}
\usepackage{stmaryrd}
\usepackage{booktabs, makecell}
\usepackage{wasysym}
\usepackage{tikzsymbols}
\usepackage{anyfontsize}
\usepackage[12pt]{moresize}
\usepackage{diagbox}

\ifdefined\isarxivversion
\usepackage[margin=1in]{geometry}
\hypersetup{colorlinks=true,citecolor=red,linkcolor=blue}
\else
\fi
\graphicspath{{./figs/}}

\def\aclpaperid{1122}

\definecolor{b2}{RGB}{51,153,255}
\definecolor{mygreen}{RGB}{80,180,0}
\definecolor{yl}{RGB}{255,80,0}

\renewcommand{\tilde}{\widetilde}
\renewcommand{\hat}{\widehat}
\newcommand{\wt}{\widetilde}

\DeclareMathOperator{\R}{{\mathbb R}}

\newcommand{\placeholder}{\texthide}
\newcommand{\texthide}{{\em TextHide}}
\newcommand{\instahide}{{\em InstaHide}}
\newcommand{\mixup}{{\em MixUp}}
\newcommand{\hideintra}{{\texthide}$_{\rm intra}$}
\newcommand{\hideinter}{{\texthide}$_{\rm inter}$}
\newcommand{\cls}{{\tt [CLS]}}
\newcommand{\sep}{{\tt [SEP]}}
\newcommand{\ETH}{{$\mathsf{ETH}$}}

\newcolumntype{P}[1]{>{\centering\arraybackslash}p{#1}}

\title{{\texthide}: Tackling Data Privacy in Language Understanding Tasks}

\date{}

\ifdefined\isarxivversion
\author{
Yangsibo Huang\thanks{\texttt{yangsibo@princeton.edu}. Princeton University.}
\and
Zhao Song\thanks{\texttt{zhaos@princeton.edu}. Princeton University and Institute for Advanced Study.}
\and
Danqi Chen\thanks{\texttt{danqic@cs.princeton.edu}. Princeton University.}
\and
Kai Li\thanks{\texttt{li@cs.princeton.edu}. Princeton University.}
\and
Sanjeev Arora\thanks{\texttt{arora@cs.princeton.edu}. Princeton University.}
}

\else
\author{Yangsibo Huang$^\dagger$ \quad Zhao Song$^\dagger$$^\ddagger$ \quad Danqi Chen$^\dagger$ \quad Kai Li$^\dagger$ \quad Sanjeev Arora$^\dagger$$^\ddagger$ \\
$^\dagger$Princeton University \quad\quad $^\ddagger$Institute for Advanced Study  \\
\texttt{\{yangsibo, zhaos\}@princeton.edu} \\
\texttt{\{danqic, li, arora\}@cs.princeton.edu}
}

\fi

\begin{document}
\ifdefined\isarxivversion

\begin{titlepage}
\maketitle
\begin{abstract}

An unsolved challenge in distributed or federated learning is to effectively mitigate privacy risks without slowing down training or reducing accuracy.  In this paper, we propose {\texthide} aiming at addressing this challenge for natural language understanding tasks.  It requires all participants to add a simple encryption step to prevent an eavesdropping attacker from recovering private text data.  Such an encryption step is efficient and only affects the task performance slightly.  In addition, {\texthide} fits well with the popular framework of fine-tuning pre-trained language models (e.g., BERT) for any sentence or sentence-pair task. We evaluate {\texthide} on the GLUE benchmark, and our experiments show that {\texthide} can effectively defend attacks on shared gradients or representations and the averaged accuracy reduction is only $1.9\%$.  We also present an analysis of the security of {\texthide} using a  conjecture about the computational intractability of a mathematical problem.\footnote{Our code is available at
\url{https://github.com/Hazelsuko07/TextHide}.}

\end{abstract}
\thispagestyle{empty}
\end{titlepage}

\else

\maketitle
\begin{abstract}

An unsolved challenge in distributed or federated learning is to effectively mitigate privacy risks without slowing down training or reducing accuracy.  In this paper, we propose {\texthide} aiming at addressing this challenge for natural language understanding tasks.  It requires all participants to add a simple encryption step to prevent an eavesdropping attacker from recovering private text data.  Such an encryption step is efficient and only affects the task performance slightly.  In addition, {\texthide} fits well with the popular framework of fine-tuning pre-trained language models (e.g., BERT) for any sentence or sentence-pair task. We evaluate {\texthide} on the GLUE benchmark, and our experiments show that {\texthide} can effectively defend attacks on shared gradients or representations and the averaged accuracy reduction is only $1.9\%$.  We also present an analysis of the security of {\texthide} using a  conjecture about the computational intractability of a mathematical problem.\footnote{Our code is available at
\url{https://github.com/Hazelsuko07/TextHide}.}

\end{abstract}

\fi


\section{Introduction}

Data privacy for deep learning has become a challenging problem for many application domains including Natural Language Processing.  For example, healthcare institutions train diagnosis systems on private patients' data~\cite{PHAM2017218,XCS2018}.  Google trains a deep learning model for next-word prediction to improve its virtual keyboard using users' mobile device data~\cite{hard2018federated}.  Such data are decentralized but moving them to a centralized location for training a model may violate regulations such as Health Insurance Portability and Accountability Act (HIPAA) \cite{act1996health} and California Consumer Privacy Act (CCPA) \cite{ccpa}.

Federated learning~\cite{mcmahan2016communication,kairouz2019advances} allows multiple parties training a global neural network model collaboratively in a distributed environment without moving data to a centralized storage. It lets each participant compute a model update (i.e., gradients) on its local data using the latest copy of the global model, and then send the update to the coordinating server. The server then aggregates these updates (typically by averaging) to construct an improved global model.

Privacy has many interpretations depending on the assumed threat models~\cite{kairouz2019advances}. This paper assumes an eavesdropping attacker with access to all information communicated by all parties, which includes the parameters of the model being trained.  With such a threat model, a recent work~\cite{zlh19} suggests that an attacker can reverse-engineer the private input.

Multi-party computation~\cite{yao82} or homomorphic encryption~\cite{g09} can ensure full privacy but they slow down computations by several orders of magnitude. Differential privacy (DP) approach~\cite{dkmmn06,d09} is another general framework to ensure certain amount of privacy by adding controlled noise to the training pipeline. However, it trades off data utility for privacy preservation.  A recent work that applies DP to deep learning was able to reduce accuracy losses~\cite{abadi2016deep} but they still remain relatively high.

The key challenge for distributed or federated learning is to ensure privacy preservation without slowing down training or reducing accuracy. In this paper, we propose {\texthide} to address this challenge for natural language understanding tasks.  The goal is to protect {\em training data privacy} at a minimal cost.  In other words, we want to ensure that an adversary eavesdropping on the communicated bits will not be able to reverse-engineer training data from any participant.

{\texthide} requires each participant in a distributed or federated learning setting to add a simple encryption step with one-time secret keys to hide the hidden representations of its text data. The key idea was inspired by {\em InstaHide}~\cite{hsla20} for computer vision tasks, which encrypts each training datapoint using a random pixel-wise mask and the {\mixup} technique~\cite{zcdl17} of data augmentation. 
However, application of {\instahide} to text data is unclear because of the well-known dissimilarities between image and language: pixel values are real numbers whereas text is sequences of discrete symbols.

{\texthide} is designed to plug into the popular framework which transforms textual input into output vectors through pre-trained language models (e.g., BERT~\cite{devlin2018bert}) and use those output representations to train a new shallow model (e.g., logistic regression) for any supervised single-sentence or sentence-pair task. The pre-trained encoder is fine-tuned as well while training the shallow model. We evaluate {\texthide} on the GLUE benchmark~\cite{wang2018glue}.  Our results show that {\texthide} can effectively defend attacks on shared gradients or representations while the averaged accuracy reduction is only $1.9\%$.

Lastly, {\texthide} and {\instahide} have completely different security arguments due to the new designs. To understand the security of the proposed approach, we also invent a new security argument using a conjecture about the computational intractability of a mathematical problem.


\vspace{-2mm}
\section{{\instahide} and Its Challenges for NLP}
\label{sec:motivation}
\vspace{-2mm}

{\instahide} \cite{hsla20} has achieved good performance in computer vision for privacy-preserving distributed learning, by providing a  cryptographic\footnote{Cryptosystem design since the 1970s seeks to ensure any attack must solves a computationally expensive task.} security while incurring much smaller utility loss and computation overhead
than the best approach based on differential privacy~\cite{abadi2016deep}.

{\instahide} is inspired by the observation that a classic computation problem, $k$-{\sc vector subset sum}\footnote{$k$-{\sc vector subset sum} is known to be hard: in the worst case, finding the secret indices requires $\geq N^{k/2}$ time \cite{al13} under the  conjecture \em{Exponential Time Hypothesis} \cite{ipz98}. See Appendix~\ref{sec:app_ksum}.}, also appears in the {\mixup}~\cite{zcdl17} method for data augmentation, which is used to improve accuracy on image data.

To encrypt an image $x \in {\mathbb R}^d$ from a private dataset, {\instahide} first picks $k-1$ other images $s_2, s_3, \ldots, s_k$ from that private dataset, or a large public dataset of $N$ images, and random nonnegative coefficients $\lambda_i$ for $i=1,.., k$ that sum to $1$, and creates a composite image $\lambda_1 x + \sum_{i=2}^k \lambda_i s_i$ ($k$ is typically small, e.g., $4$). A composite label is also created using the same set of coefficients.\footnote{Only the labels of the examples from the private dataset will get combined. See \citep{hsla20} or Section~\ref{sec:texthide} for more details.}
Then it adds another layer of security: pick a {\em random mask}  $\sigma \in \{-1, 1\}^d$ and output the encryption $\tilde{x} = \sigma\circ (\lambda_1 x + \sum_{i=2}^k \lambda_i s_i)$, where $\circ$ is coordinate-wise multiplication of vectors.
The neural network is then trained on encrypted images, which look like random pixel vectors to the human eye and yet lead to  good classification accuracy.
Note that the \textquotedblleft one-time secret key\textquotedblright\  $\sigma, s_2, \cdots, s_k$ used to encrypt $x$ will not be reused to encrypt other images.

\paragraph{Challenges of applying {\instahide} to NLP.}
There are two challenges to apply {\instahide} to text data for language understanding tasks. The first is the discrete nature of text, while the encryption in {\instahide} operates at continuous inputs. The second is that most NLP tasks today are solved by \textit{fine-tuning} pretrained language models such as BERT on downstream tasks. It remains an open question how to add encryption into such a framework and what type of security argument it will provide. The following section presents our approach that overcomes these two challenges.

\begin{figure*}[t]
    \centering
    \includegraphics[width = 0.98 \textwidth]{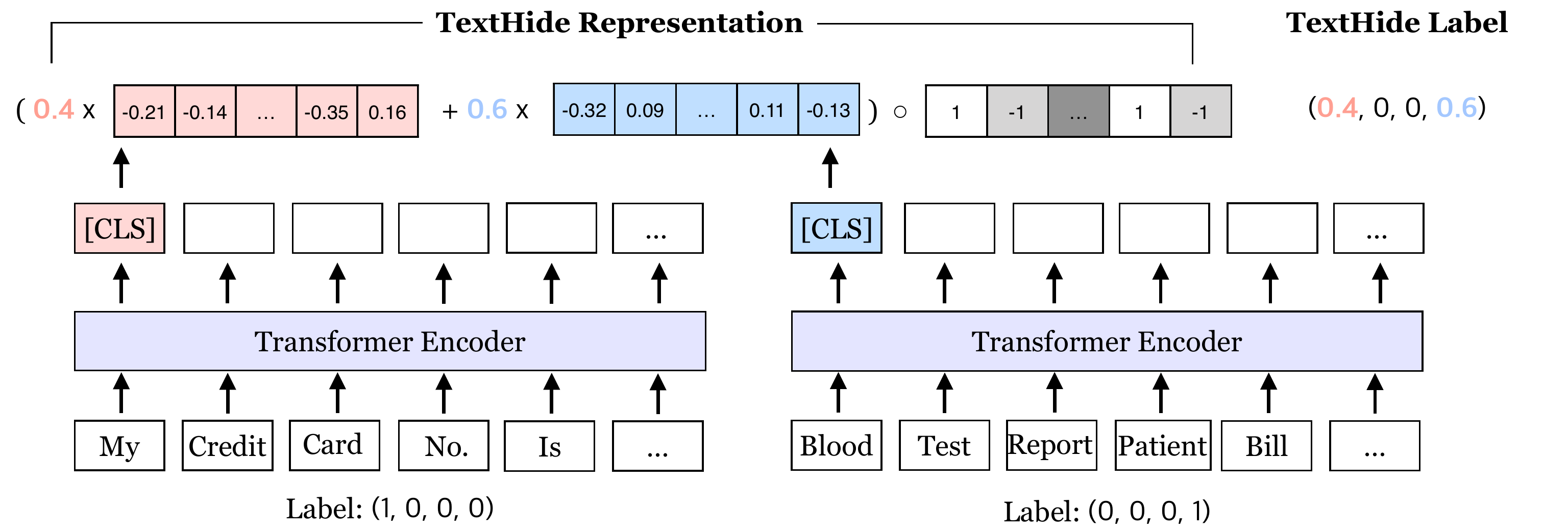}
    \caption{\small An illustration of {\texthide} encryption with $k=2$, where $k$ is the number of inputs (sentence or sentence-pair) got mixed in each {\texthide} representation. {\texthide} first encodes each text input using a transformer encoder, then linearly combines their output representations (i.e., {\cls} tokens), as well as their labels. Finally, an entry-wise mask is chosen from a randomly pre-generated pool and applied on the mixed representation. The entry-wise mask, together with the other datapoints to mix constitute the ``one-time secret key'' of the {\texthide} scheme. Note that training directly takes place on encrypted data and no decryption is needed.
    }
    \label{fig:texthide}
    \vspace{-6mm}
\end{figure*}


\section{{\texthide}: Formal Description}
\label{sec:texthide}

There are two key ideas in {\texthide}. The first one is using the ``one-time secret key'' coming from {\instahide} for encryption, and the second  is a method to incorporate such encryption into the popular framework of fine-tuning a pre-trained language model e.g., BERT~\cite{devlin2018bert}.

In the following, we will describe how to integrate {\texthide} in the federated learning setting~(Section~ \ref{sec:finetune}), and then present two {\texthide} schemes (Section~\ref{sec:coordinate_wise} and \ref{sec:cross_dataset}). We analyze the security of {\texthide} in Section~\ref{sec:app_sec}.

\subsection{Fine-tuning BERT with {\texthide}}
\label{sec:finetune}

In a federated learning setting, multiple participants holding private text data may wish to solve NLP tasks by using a BERT-style fine-tuning pipeline, where {\texthide}, a simple {\instahide}-inspired encryption step can be applied at its intermediate level to ensure privacy (see Figure~\ref{fig:texthide}).

The BERT fine-tuning framework assumes (input, label) pairs $(x, y)$'s,  where $x$ takes the form of \cls $s_1$ \sep for single-sentence tasks, or \cls $s_1$ \sep $s_2$ \sep for sentence-pair tasks. $y$ is a one-hot vector for classification tasks, or a real-valued number for regression tasks.\footnote{We will mainly use classification tasks as examples throughout the paper for brevity.} For a standard fine-tuning process, federated learning participants use a BERT-style model $f_{\theta_1}$ to compute hidden representations $f_{\theta_1}(x)$'s for their inputs $x$'s and then train a shallow classifier $h_{\theta_2}$ on $f_{\theta_1}(x)$, while also fine-tuning $\theta_1$. The parameter vectors $\theta_1, \theta_2$ will be updated at the central server via pooled gradients. All participants hold current copies of the two models.

To ensure privacy of their individual inputs $x$'s, federated learning participants can apply {\texthide} encryption at the {\em output} $f_{\theta_1}(x)$'s. The model $h_{\theta_2}$ will be trained on these encrypted representations. Each participant will compute gradients by backpropagating through their private encryption, and this is going to be the source of the secrecy: the attacker can see the communicated gradients but not the secret encryptions, which limits leakage of information about the input. 

We then formally describe two {\texthide} schemes for fine-tuning BERT in the federated learning setting: {\texthide}$_{\rm intra}$ which encrypts an input using other examples from the same dataset, and {\texthide}$_{\rm inter}$ which utilizes a large public dataset to perform encryption. 
Due to a large public dataset, {\texthide}$_{\rm inter}$ is more secure than {\texthide}$_{\rm intra}$, but the latter is quite secure in practice when the training set is large.

\subsection{Basic {\texthide}: Intra-Dataset {\texthide}}
\label{sec:coordinate_wise}

In {\texthide}, we have a pre-trained text encoder $f_{\theta_1}$, which takes $x$, a sentence or a sentence pair, and maps it to a representation $e=f_{\theta_1}(x)\in \R^d$ (e.g., $d=768$ for BERT$_{\rm base}$). We use $[b]$ to denote the set $\{1,2,\cdots,b\}$. Given a dataset ${\cal D}$, we denote the set $\{x_i, y_i\}_{i \in [b]}$ an ``input batch'' by $\mathcal{B}$, where $x_1, \cdots, x_b$ are $b$ inputs randomly drawn from ${\cal D}$, and $y_1, \cdots, y_b$ are their labels. For each $x_i$ in the batch $\mathcal{B}$, $i \in [b]$, we can encode $x_i$ using $f_{\theta_1}$, and obtain a new set of  $\{e_i = f_{\theta_1} (x_i), y_i\}_{i \in [b]}$. We refer to this set as an ``encoding batch'', and denote it by $\mathcal{E}$. Later in this section, we use $\tilde{e}_i$ to denote the {\texthide} encryption of $e_i$ for $i \in [b]$, and name the set ${\tilde{\mathcal{E}}}=\{\tilde{e}_i, y_i\}_{i \in [b]}$ as a ``hidden batch'' of $\mathcal{E}$.

\begin{algorithm}[t]
\caption{\small$(m,k)$-{\texthide}}
\label{alg:texthide}
\small
\begin{algorithmic}[1]
\Procedure{\textsc{TextHide}}{$ \mathcal{E}, {\cal M}, k$}
\State \Comment{$\mathcal{E}$: the training batch, $b$: $|\mathcal{E}|=b$}
\State \Comment{${\cal M}$: the mask pool, $m$: $|{\cal M}| = m$}
\State \Comment{$k$: number of training examples to be mixed}
\State \Comment{Let $[b]$ denote the set $\{1,2,\cdots,b\}$}
    \State $\wt{\mathcal{E}} \leftarrow \emptyset$
     \State Generate $\pi_1$ such that $\pi_1(i)=i$, $\forall i \in [b]$
    \State Generate $k-1$ random permutations $\pi_2$, $\cdots$, $\pi_k : [b] \rightarrow [b]$
    \State Sample $\lambda_1, \cdots, \lambda_b \sim |{\cal N}(0,I_k)| \in \R^k$ uniformly at random, normalize s.t. $\sum_{j=1}^k (\lambda_i)_{j} = 1$, $\forall i \in [b]$.
    \For{$(e_i, y_i) \in  \mathcal{E} $}
        \State $\sigma_i \sim {\cal M}$
        \State $\tilde{e}_i \leftarrow \sigma_i \circ \sum_{j=1}^{k} (\lambda_i)_j \cdot e_{ \pi_j( i ) }$ \label{line:mix_x}
        \State $\tilde{y}_i \leftarrow \sum_{j=1}^{k} (\lambda_{i})_j \cdot  y_{\pi_j( i )}$ \label{line:mix_label}
    \State $\wt{\mathcal{E}} \leftarrow \wt{\mathcal{E}} \cup \{ (\wt{x}_i , \wt{y}_i ) \}$
    \EndFor
    \State \Return $\tilde{\mathcal{E}}$
\EndProcedure
\end{algorithmic}
\end{algorithm}

\begin{algorithm}[t]
\small
\caption{\small Federated fine-tuning BERT using $(m,k)$-{\texthide} with $C$ clients (indexed by $c$)}
\label{alg:bert_texthide}
\begin{algorithmic}[1]
\State {$m$: size of each client's mask pool}
\State {$k$: number of training samples to be mixed}
\State {$d$: hidden size (e.g., 768 in BERT)}

\Procedure{ServerExecution}{$f_{\theta_1}, h_{\theta_2}$}
\State \Comment{$f_{\theta_1}$: the pre-trained BERT; $h_{\theta_2}$: a shallow classifer}
\State \Comment{$T$: number of model updates, $\eta$: learning rate}
\State $f_{\theta_{1}^1} \leftarrow f_{\theta_1}$, $h_{\theta_{2}^1} \leftarrow h_{\theta_2}$
\For{$t=1 \to T$}
\For{each client $c$ {\bf in parallel}}
    \State $\nabla_{\theta_1^t, c},\nabla_{\theta_2^t, c} \leftarrow {\text{\sc CliUpdate}}(c, f_{\theta^t_{1}},h_{\theta^t_{2}})$
\EndFor
\State $\theta_{1}^{t+1} \gets \theta_{1}^{t} - \frac{\eta}{C} \sum_{c=1}^C  \nabla_{\theta_1^t, c}$ \label{line:global_update1}
\State $\theta_{2}^{t+1} \gets \theta_{2}^{t} - \frac{\eta}{C} \sum_{c=1}^C  \nabla_{\theta_2^t, c}$ \label{line:global_update2}
\EndFor
\State \Return $f_{\theta_1}^{T+1}, h_{\theta_2}^{T+1}$
\EndProcedure
\Procedure{CliUpdate}{$c, f_{\theta_{1}}, h_{\theta_{2}}$} \Comment{Run on Client $c$}
\State \Comment{$b$: batch size; ${\cal D}^c$: private train set of client $c$}
\State \Comment{${\cal M}_c$: the mask pool of size $m$ owned by client $c$, masks are sampled i.i.d. from $\{-1,+1\}^d$ }

\State Sample a random batch $ \{x_{i}, y_{i}\}_{i \in [b]} $ from $\mathcal{D}_c$
\State $\mathcal{E} = \{  f_{\theta_1} (x_i), y_i \}_{i \in [b]}$ \label{line:get_rep}
\State $\tilde{\mathcal{E}} \leftarrow \textsc{{\texthide}}(\mathcal{E}, \mathcal{M}_c, k)$
\label{line:hide_batch}
\State \Return $\nabla_{\theta_1} \mathcal{L}(f_{\theta_{1}}, h_{\theta_{2}}; \tilde{\mathcal{E}})$, $\nabla_{\theta_2} \mathcal{L}(f_{\theta_{1}}, h_{\theta_{2}}; \tilde{\mathcal{E}})$
\label{line:local_update}
\EndProcedure
\end{algorithmic}
\end{algorithm}
\setlength{\textfloatsep}{10pt}

We use $\sigma \in \{-1, +1\}^d$ to denote an entry-wise sign-flipping mask. For a {\texthide} scheme, ${\cal M} = \{\sigma_1, \cdots, \sigma_m \}$ denotes its randomly pre-generated mask pool of size $m$, and $k$ denotes the number of sentences combined in a {\texthide} representation. We name such a parametrized scheme as $(m,k)$-{\texthide}.

\paragraph{$(m,k)$-{\texthide}.} Algorithm~\ref{alg:texthide} describes how $(m,k)$-{\texthide} encrypts an encoding batch $\mathcal{E}=\{e_i, y_i\}_{i \in [b]}$ into a hidden batch $\tilde{\mathcal{E}}$, where $b$ is the batch size. For each $e_i$ in $\mathcal{E}$, {\texthide} linearly combines it with $k-1$ other representations, as well as their labels. Then, {\texthide} randomly selects a mask $\sigma_i$ from ${\cal M}$, the mask pool, and applies it on the combination using coordinate-wise multiplication. This gives $\tilde{e}_i$, the encryption of $e_i$ (lines \ref{line:mix_x}, \ref{line:mix_label} in Algorithm~\ref{alg:texthide}).
Note that different $e_i$'s in the batch get assigned to a fresh random $\sigma_i$'s from the pool.

\paragraph{Plug into federated BERT fine-tuning.}
Algorithm~\ref{alg:bert_texthide} shows how to incorporate {$(m,k)$-\texthide} in federated learning, to allow a centralized server and $C$ distributed clients collaboratively fine-tune a language model (e.g., BERT) for any downstream tasks, without sharing raw data. Each client (indexed by $c$) holds its own private data ${\cal D}_c$ and a private mask pool ${\cal M}_c$, and $\sum_{c=1}^C |{\cal M}_c| = m$.

The procedure takes a pre-trained BERT $f_{\theta_1}$ and an initialized task-specific classifier $h_{\theta_2}$, and runs $T$ steps of {\em global} updates of both $\theta_1$ and $\theta_2$. In each {\em global} update, the server aggregates {\em local} updates of $C$ clients. For a {\em local} update at client $c$, the client receives the latest copy of $f_{\theta_1}$ and $h_{\theta_2}$ from the server, samples a random input batch $\{x_i, y_i\}_{i \in [b]}$ from its private dataset ${\cal D}_c$, and encodes it into an encoding batch $\mathcal{E} = \{  e_i=f_{\theta_1} (x_i), y_i \}_{i \in [b]}$ (line \ref{line:get_rep} in Algorithm~\ref{alg:bert_texthide}).

To protect privacy, each client will run $(m,k)$-{\texthide} with its own mask pool ${\cal M}_c$ to encrypt the encoding batch $\mathcal{E}$ into a hidden batch $\tilde{\mathcal{E}}$ (line \ref{line:hide_batch} in Algorithm~\ref{alg:bert_texthide}). The client then uses the hidden batch $\tilde{\mathcal{E}}$ to calculate the model updates (i.e., gradients) of both the BERT encoder $f_{\theta_1}$ and the shallow classifier $h_{\theta_2}$, and returns them to the server (line \ref{line:local_update} in Algorithm~\ref{alg:bert_texthide}). The server averages all updates from $C$ clients, and runs a {\em global} update for $f_{\theta_1}$ and $h_{\theta_2}$ (lines \ref{line:global_update1}, \ref{line:global_update2} in Algorithm~\ref{alg:bert_texthide}).

\subsection{Inter-dataset {\texthide}}
\label{sec:cross_dataset}
Inter-dataset {\texthide} encrypts private inputs with text data from a {\em{second}} dataset, which can be a large public corpus (e.g., Wikipedia). The large public corpus plays a role reminiscent of the {\em random oracle} in cryptographic schemes~\cite{canetti2004random}. 

Assume we have a private dataset $D_{\text{private}}$ and a large public dataset $D_{\text{public}}$, {\hideinter} randomly chooses $\lceil k/2 \rceil$ sentences from $D_{\text{private}}$
and the other $\lfloor k/2 \rfloor$ from $D_{\text{public}}$, mixes their representations, and applies on it a random mask from the pool. A main difference between {\hideinter} and {\hideintra} is, {\hideintra} mixes all labels of inputs used in the combination, while in {\hideinter}, only the labels from $D_{\text{private}}$ will be mixed (there is usually no label from the public dataset).
Specifically, for an original datapoint $\{x_i, y_i\} \in {\cal E}$, let $S \subset [b]$ denote the set of data points' indices that its {\texthide} encryption combines, and $|S| = k$. Then its {\hideinter} label is given by
\begin{align*}
\frac{ \sum_{j = 1}^k (\lambda_i)_j \cdot y_{ \pi_j( i ) } \cdot {\bf 1} [ \pi_j( i) \in {D_{\text{private}} \cap S} ] }{ \sum_{j=1}^k (\lambda_i)_j \cdot {\bf 1} [ \pi_j( i) \in {D_{\text{private}} \cap S} ] },
\end{align*}
where ${\bf 1}[f]$ is a variable that ${\bf 1}[f] = 1$ if $f$ holds, and $=0$ otherwise. For each $j \in [k]$, $\pi_j : [b] \rightarrow [b]$ is a permutation.


\subsection{On Security of \texthide}
\label{sec:app_sec}
The encrypted representations produced by \texthide\ themselves are secure --- i.e., do not allow any efficient way to recover the text $x$ --- from the security framework of {\instahide} (see Appendix~\ref{sec:app_ksum} for $k$-{\sc vector subset sum}). However, an additional source of information leakage is the shared gradients during federated learning, as shown by~\cite{zlh19}. We mitigate this by ensuring that the secret mask $\sigma$ used to encrypt the representation of input $x$ is changed each epoch. The pool of masks is usually much larger than the number of epochs, which means that each mask gets used only once for an input (with negligible failure probability). The gradient-matching attack of ~\cite{zlh19} cannot work in this scenario. In the following section, we will show that it does not even work with a fixed mask.
\section{Experiments}
\label{sec:exp}

We evaluate the utility and privacy of {\texthide} in our experiments.  We aim to answer the following questions in our experiments:
\vspace{-2mm}
\begin{itemize}
    \item What is the accuracy when using {\texthide} for sentence-level natural language understanding tasks (Section~\ref{sec:utility})?
    \vspace{-2mm}
    \item How effective is {\texthide} in terms of hiding the gradients (Section~\ref{sec:privacy_grad}) and the representations of the original input (Section~\ref{sec:privacy_embed})?
\end{itemize}
\vspace{-3mm}

\begin{table*}[t]
\setlength{\tabcolsep}{1.5pt}
    \centering
    \begin{tabular}{lccccccccc}
    \toprule
    {\bf Dataset} & $|\cal D|$ & {\bf Task} & {\bf Metric} & {\bf Baseline} & {\bf \hideintra} & {\bf \hideinter}\\
    \midrule
    {RTE} &  2.5k & NLI & Acc. & $72.0_{(0.86)}$ & $65.2_{(1.71)}$ & $54.4_{(1.82)}$ \\
    {MRPC} & 3.7k & Paraphrase &  F1 / Acc. & $90.2_{(0.80)}$ / $86.2_{(1.40)}$ & $89.7_{(0.56)}$ / $85.6_{(0.96)}$ & $88.1_{(0.52)}$ / $82.6_{(0.75)}$\\
    {STS-B} &  7k & Similarity & P / S corr. & $90.1_{(0.12)}$ / $89.7_{(0.17)}$ & $87.0_{(0.25)}$ / $87.0_{(0.27)}$ & $86.0_{(0.27)}$ / $86.2_{(0.19)}$\\
    {CoLA} &  8.5k & Acceptability & MCC & $58.9_{(1.00)}$ & $56.3_{(0.86)}$ & $52.3_{(0.80)}$\\
    {SST-2} & 67k & Sentiment & Acc. & $92.4_{(0.76)}$ & $91.7_{(0.51)}$ & $91.3_{(0.41)}$\\
    {QNLI} & 108k & NLI & Acc. & $91.7_{(0.70)}$ & $91.0_{(0.31)}$ & $89.8_{(0.56)}$\\
    {QQP} & 364k & Paraphrase & F1 / Acc. & $87.9_{(0.39)}$ / $91.0_{(0.30)}$ & $87.3_{(0.41)}$ / $90.5_{(0.33)}$ & $86.5_{(0.28)}$ / $89.8_{(0.14)}$\\
    {MNLI} & 393k & NLI & m/mm & $86.1_{(0.36)}$ / $85.6_{(0.23)}$ & $84.0_{(0.15)}$ / $84.1_{(0.23)}$ & -\\
    \bottomrule
    \end{tabular}
    \caption{Performance on the GLUE tasks for both baseline (standard finetuning) and {\texthide} with BERT$_{\rm base}$, measured on the development sets. We report the mean results across 5 runs, with  $(m,k)=(16,4)$ for RTE and $(m,k)=(256,4)$ for all the other datasets (see text for more details). Standard deviations are reported in parentheses. $|\cal D|$ denotes the number of training examples.  {\texthide} only suffers minor utility loss: $< 3\%$ in most cases for both \hideintra and \hideinter. 
    `P / S corr.' is Pearson/Spearman correlation and `MCC' is Matthew's correlation.
    }
    \vspace{-4mm}
    \label{tab:summary_exp}
\end{table*}

\subsection{Experimental Setup}
\label{sec:setup}
\paragraph{Dataset.}
We evaluate {\texthide} on the General Language Understanding
Evaluation (GLUE) benchmark~\cite{wang2018glue}, a collection of 9 sentence-level language understanding tasks:

\begin{itemize}
    \vspace{-3mm}
    \item Two {\em sentence-level classification} tasks including Corpus of Linguistic Acceptability (CoLA)~\cite{warstadt2018neural}, and Stanford Sentiment Treebank (SST-2)~\cite{socher2013recursive}.
    \vspace{-3mm}
    \item Three {\em sentence-pair similarity} tasks including Microsoft Research Paraphrase Corpus (MRPC)~\cite{dolan2005automatically}, Semantic Textual Similarity Benchmark (STS-B)~\cite{cer2017semeval}, and Quora Question Pairs (QQP)\footnote{\href{https://www.quora.com/q/quoradata/First-Quora-Dataset-Release-Question-Pairs}{https://www.quora.com/q/quoradata/First-Quora-Dataset-Release-Question-Pairs}}.
    \vspace{-2mm}
    \item Four {\em natural language inference}~(NLI) tasks including Multi NLI (MNLI)~\cite{williams2018broad}, Question NLI (QNLI)~\cite{rajpurkar2016squad}, Recognizing Textual Entailment (RTE)~\cite{dagan2006pascal, bar2006second, giampiccolo2007third}, and Winograd NLI (WNLI)~\cite{levesque2011winograd}.
    \vspace{-2mm}
\end{itemize}

Following previous work \cite{devlin2018bert, joshi2020spanbert}, we exclude WNLI in the evaluation.
Table~\ref{tab:summary_exp} summarizes the data size, tasks and evaluation metrics of all the datasets. All tasks are single-sentence or sentence-pair classification tasks except that STS-B is a regression task.

\paragraph{Implementation.} We fine-tune the pre-trained cased  BERT$_{\rm base}$ model released by \cite{devlin2018bert} on each dataset. We notice that generalizing to different masks requires a more expressive classifier, thus instead of adding a linear classifier on top of the {\cls} token, we use a multilayer perceptron of hidden-layer size (768, 768, 768) to get better performance under {\texthide}.   We use AdamW~\cite{kingma2014adam} as the optimizer, and a linear scheduler with a warmup ratio of 0.1. More details of hyperparameter selection are given in~Appendix~\ref{sec:app_hyperparam}. To show {\texthide}'s compatibility with the state-of-the-art model, we also test with the RoBERTa$_{\rm base}$ model released by~\cite{liu2019roberta} and report the results in Appendix~\ref{sec:app_more_exp}.

\begin{figure*}[!ht]
    \centering
    \subfloat[RTE ($|\cal D|$: 2.5k)]{\includegraphics[width=0.245\linewidth]{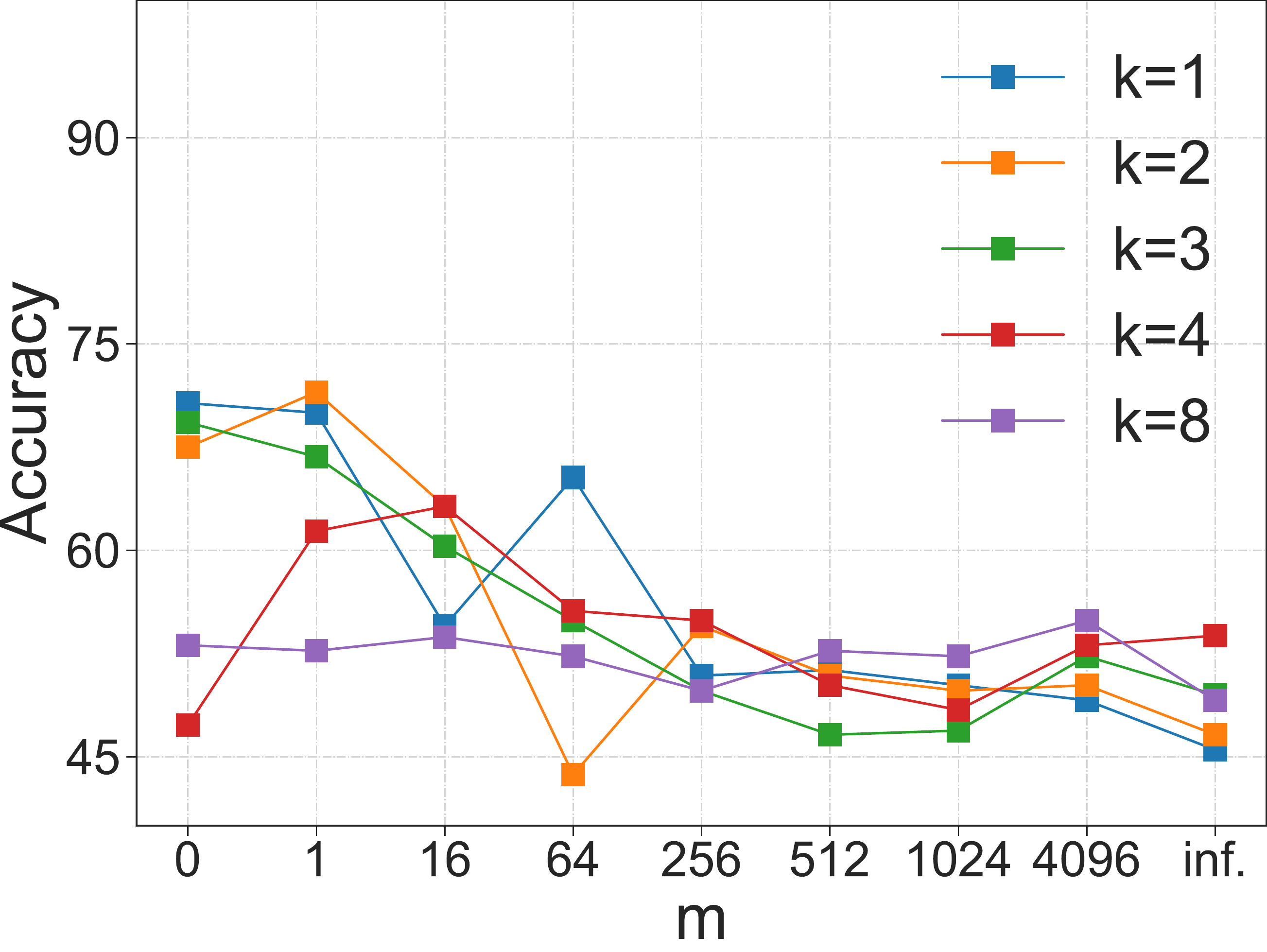}}
    \hspace{0mm}
    \subfloat[MRPC ($|\cal D|$: 3.7k)]{\includegraphics[width=0.245\linewidth]{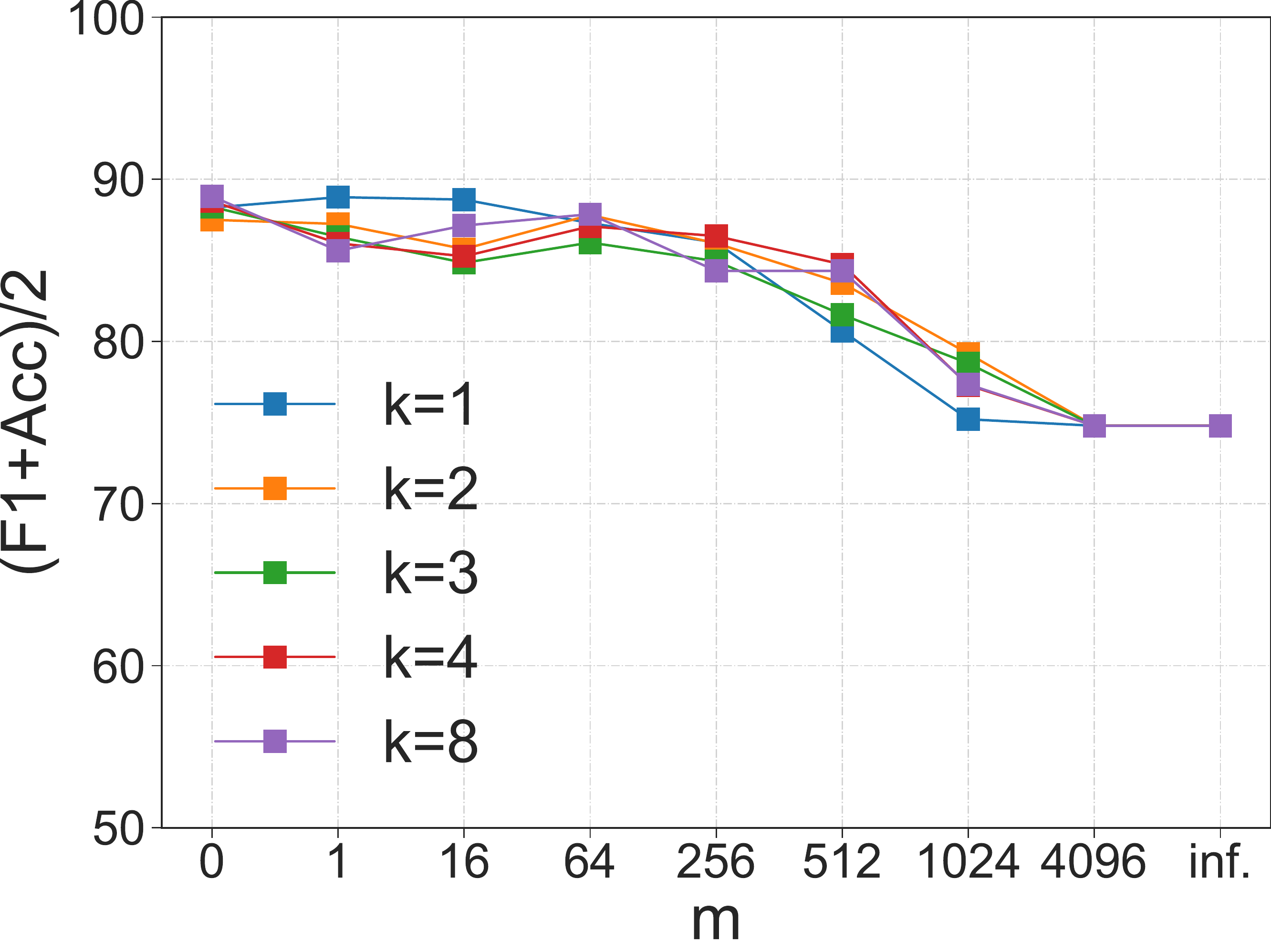}}
    \hspace{0mm}
    \subfloat[STS-B ($|\cal D|$: 7k)]{\includegraphics[width=0.245\linewidth]{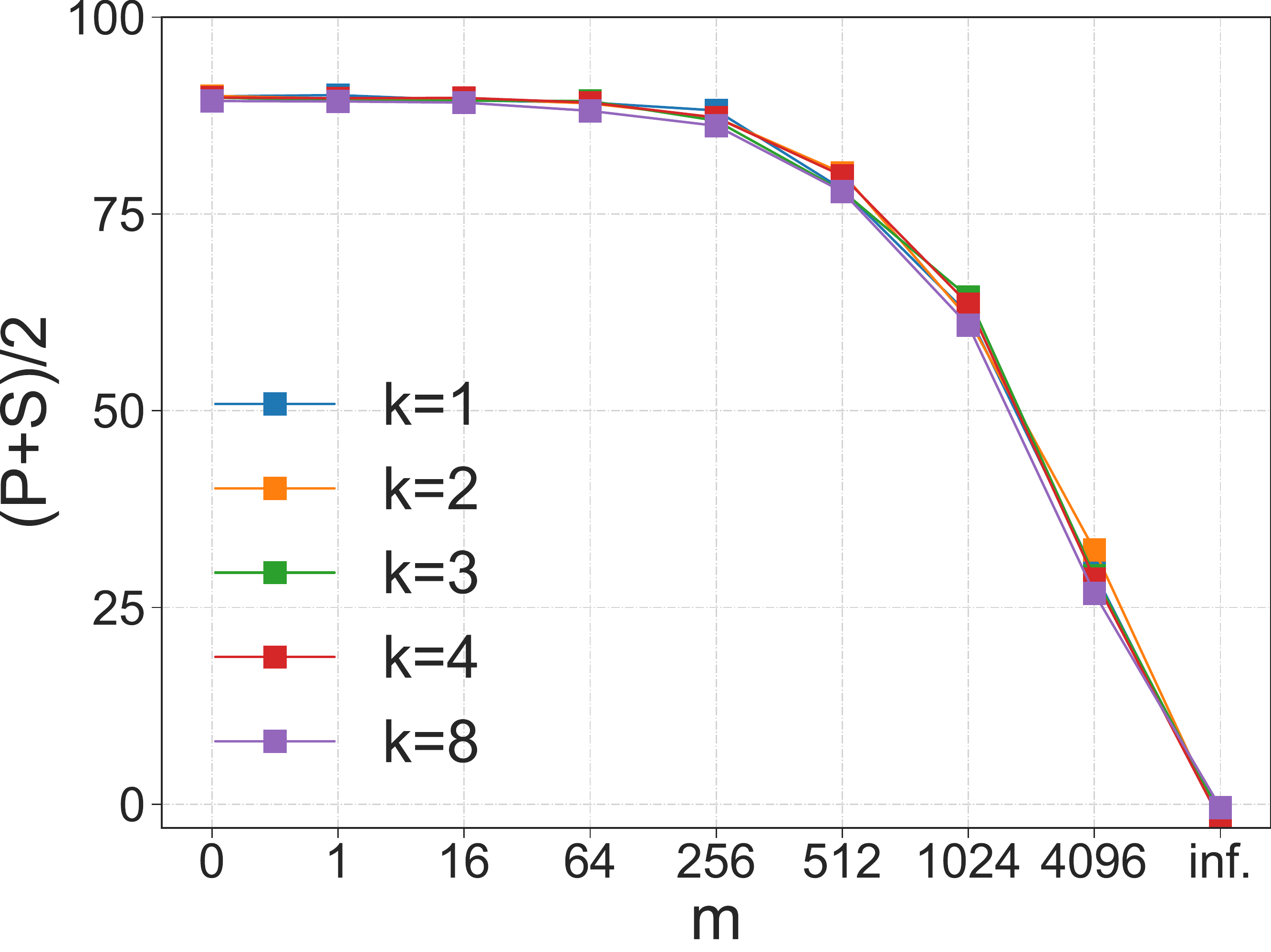}}
    \hspace{0mm}
    \subfloat[CoLA ($|\cal D|$: 8.5k)]{\includegraphics[width=0.245\linewidth]{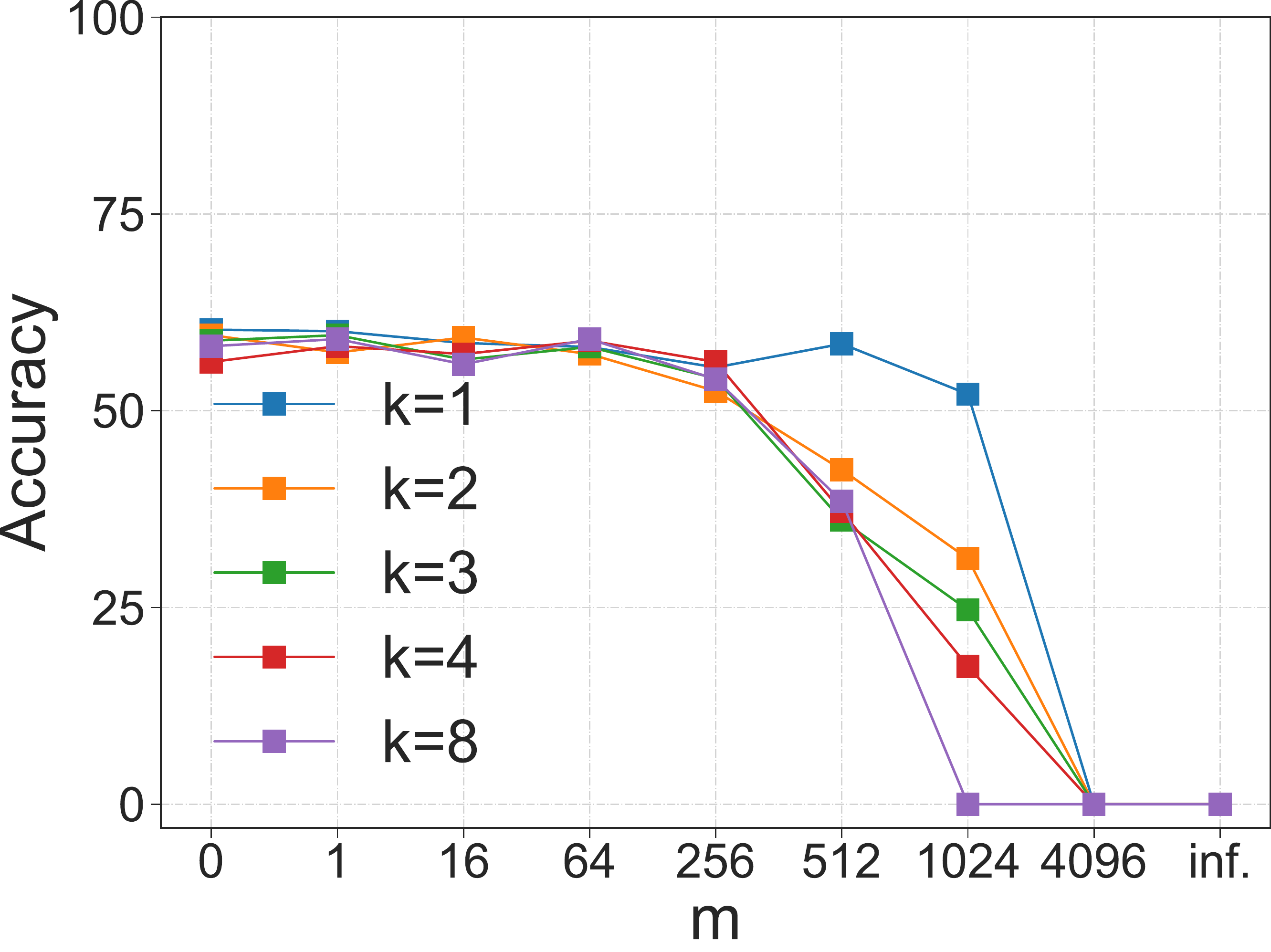}} \\
    \vspace{-3mm}
    \subfloat[SST-2 ($|\cal D|$: 67k)]{\includegraphics[width=0.245\linewidth]{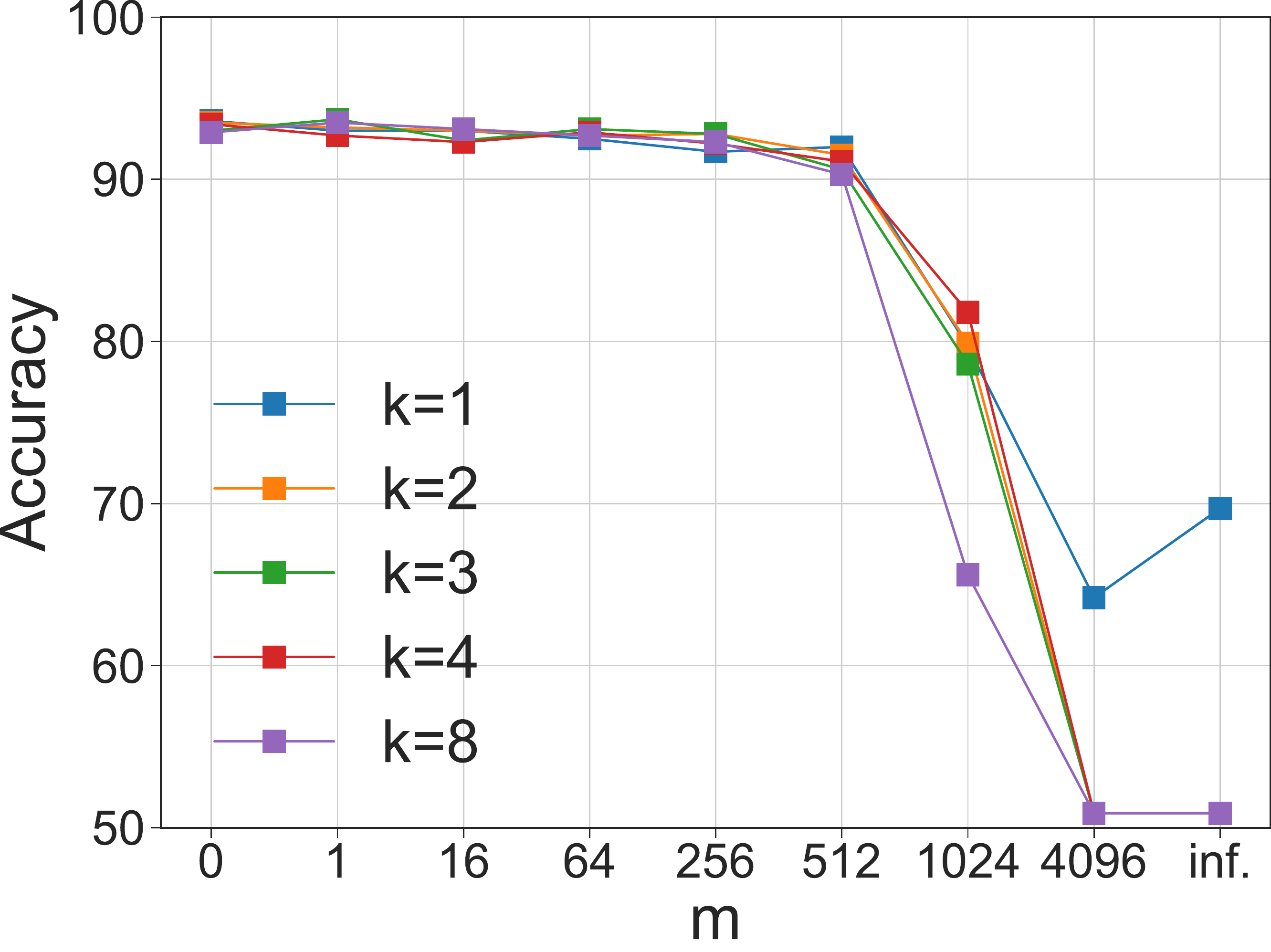}}
    \hspace{0mm}
    \subfloat[QNLI ($|\cal D|$: 108k)]{\includegraphics[width=0.245\linewidth]{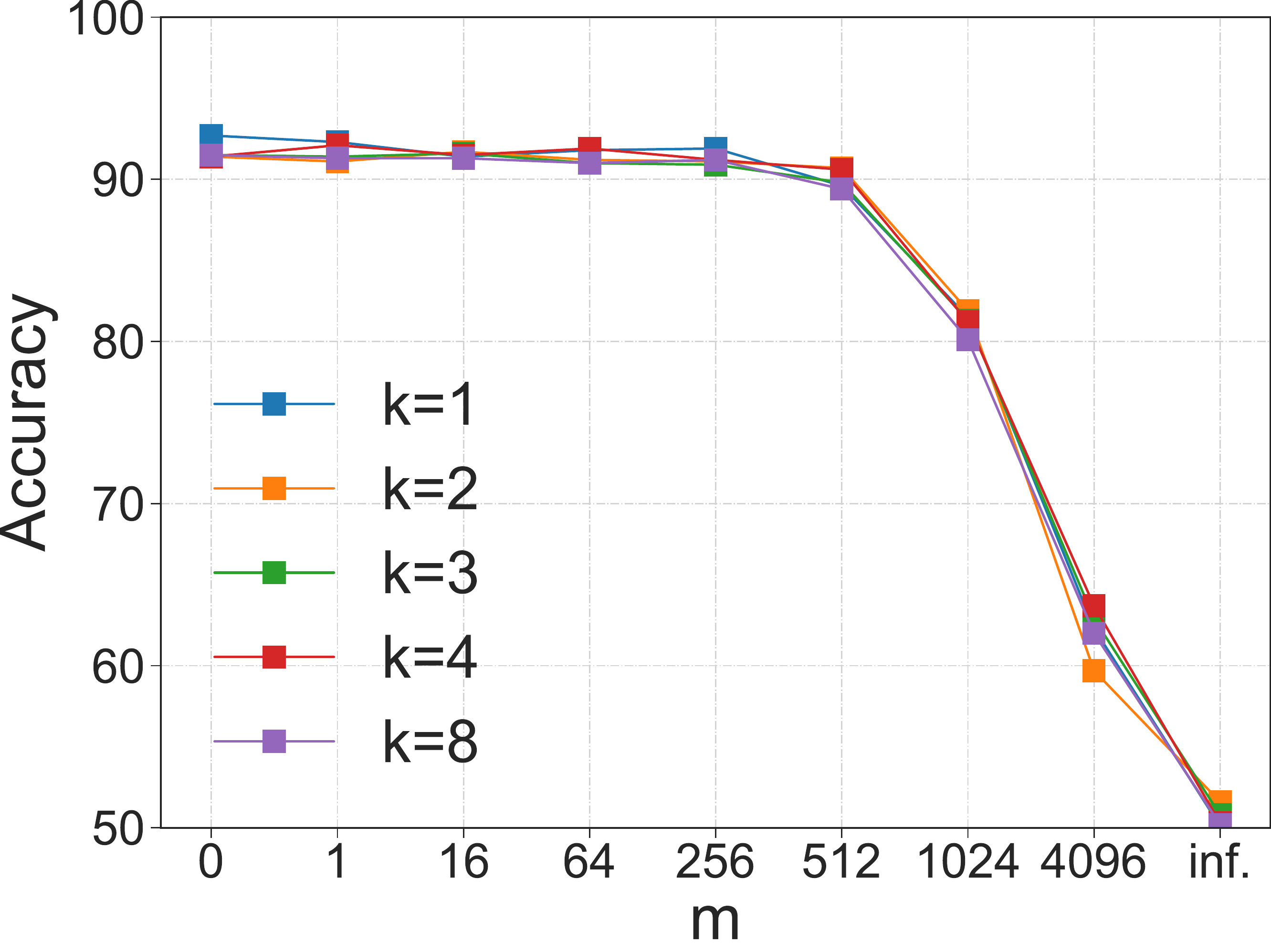}}
    \hspace{0mm}
    \subfloat[QQP ($|\cal D|$: 364k)]{\includegraphics[width=0.245\linewidth]{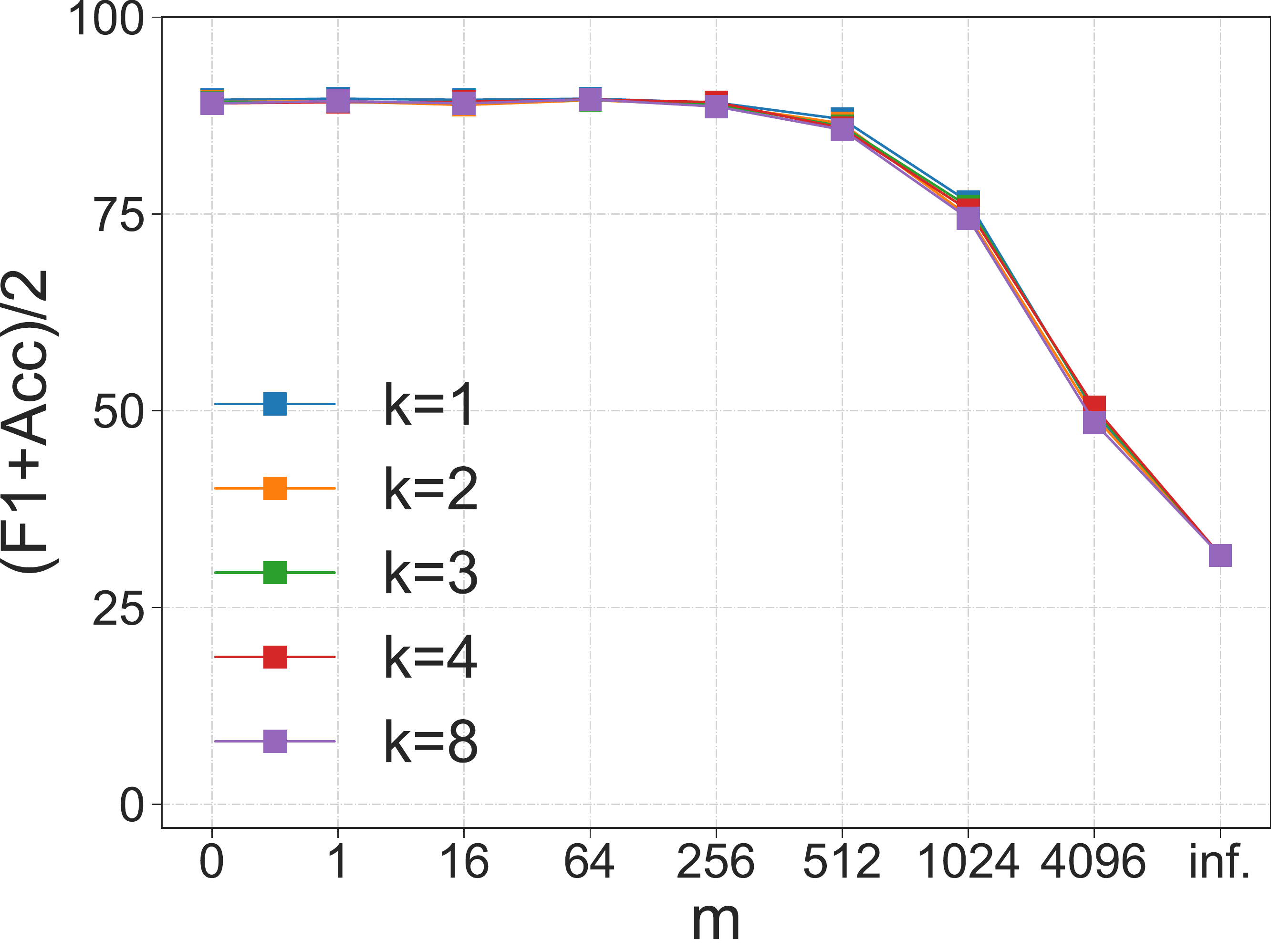}}
    \hspace{0mm}
    \subfloat[MNLI ($|\cal D|$: 393k)]{\includegraphics[width=0.245\linewidth]{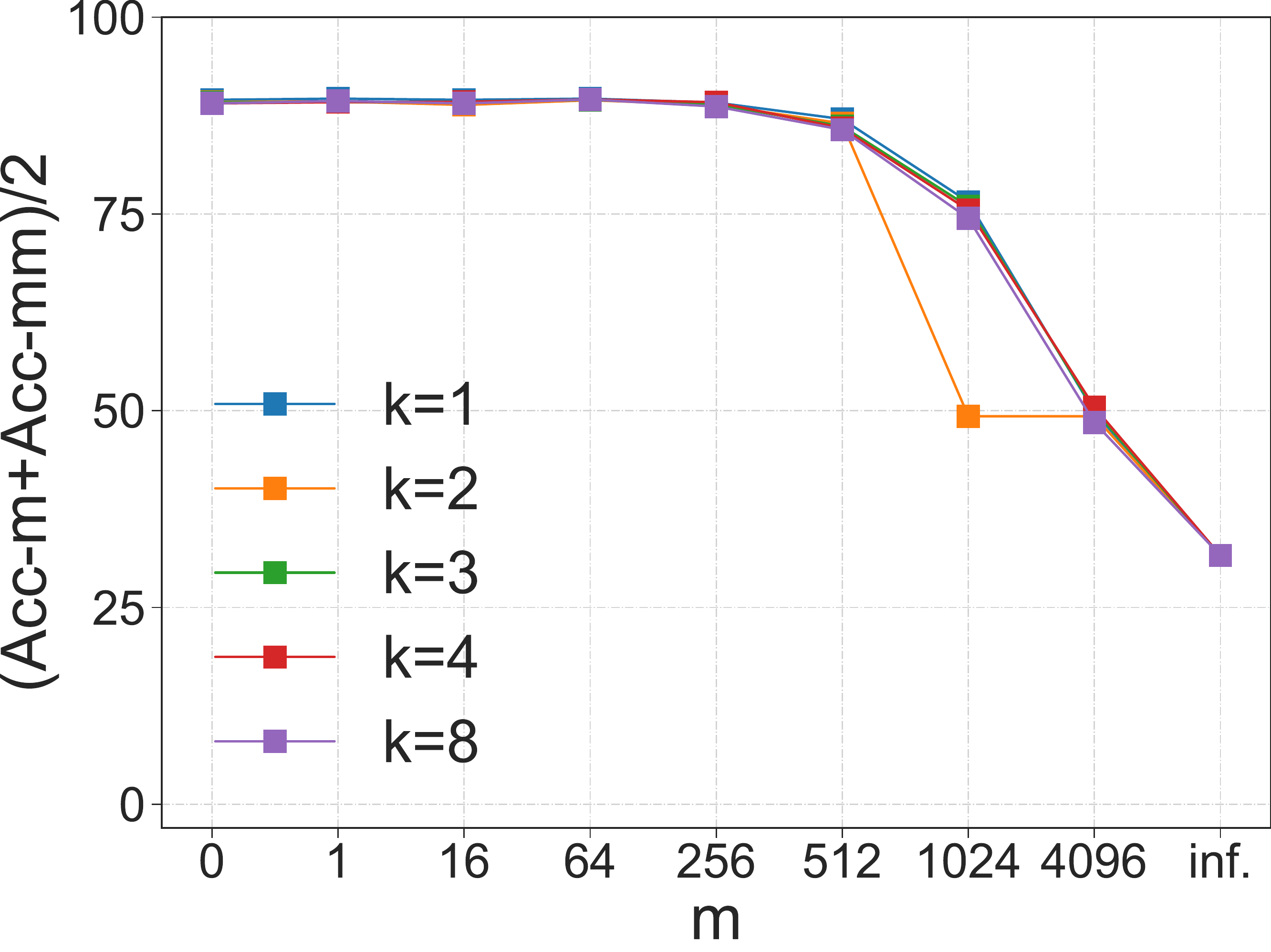}}
    \caption{Performance of {\hideintra} on the GLUE tasks of different $(m,k)$ pairs, measured on the development sets. $(m,k)=(0,1)$ is equivalent to the baseline. Metrics are marked on the y-axis. $|\cal D|$ denotes the number of training examples.
    {\texthide} with $m=256$ achieves good utility on all datasets (except RTE). Larger dataset can work with larger $m$. 
    \vspace{-6mm}
    }
    \label{fig:km_vis}
\end{figure*}

\subsection{Accuracy Results of {\texthide}}
\label{sec:utility}
To answer the first question, we compare the accuracy of {\texthide} to the BERT baseline without any encryption. 

We the vary {\texthide} scheme as follows:
\begin{itemize}
    \vspace{-2mm}
    \item Evaluate different $(m, k)$ combinations, where $m$ (the size of mask pool) is chosen from $\{0, 1, 16, 64, 256, 512, 1024, 4096, \infty\}$, and $k$ (the number of inputs to combine) is chosen from $\{1, 2, 3, 4, 8\}$. $(m,k) = (0, 1)$ is equivalent to the baseline.
    \vspace{-2mm}
    \item Test both {\hideintra} and {\hideinter}. We use MNLI train set (around 393k examples and all the labels are removed) as the ``public dataset'' in the inter-dataset setting and run BERT fine-tuning with {\hideinter} on the other 7 datasets. Here we use MNLI simply for convenience as it is the largest dataset in GLUE and one can use any public corpora (e.g., Wikipedia) in principle.
\end{itemize}

\paragraph{Results with different $(m,k)$ pairs.} Figure~\ref{fig:km_vis} shows the performance of {\hideintra} parameterized with different $(m,k)$'s. When $m$ is fixed, the network performs consistently with different $k$'s, suggesting that {\mixup}~\cite{zcdl17} also works for language understanding tasks.

Increasing $m$ makes learning harder since the network needs to generalize to different masking patterns. However, for most datasets (except for RTE), {\texthide} with $m=256$ only reduces accuracy slightly comparable to the baseline. Our explanation for the poor performance on RTE is that we find training on this small dataset (even without encryption) to be quite unstable. This has been observed in~\cite{dodge2020fine} before.
In general, {\texthide} can work with larger $m$ (better security) when the training corpus is larger (e.g., $m=512$ for data size $> 100$k).

\paragraph{{\hideintra} vs. {\hideinter}.}
{\hideintra} mixes the representations from the same private dataset, whereas {\hideinter} combines representations of private inputs with representations of random inputs from a large public corpus (MNLI in our case).

Table~\ref{tab:summary_exp} shows the results of the baseline and {\texthide} (both {\hideintra} and {\hideinter}) on the GLUE benchmark, with $(m,k)=(256,4)$ except for RTE with $(m,k)=(16,4)$. The averaged accuracy reduction of {\hideintra} is $1.9\%$, when compared to the baseline model. With the same $(m,k)$, {\hideinter} incurs an additional $2.5\%$ accuracy loss on average, but as previously suggested, the large public corpus gives a  stronger notion of security.
\subsection{Security of Gradients in {\texthide}}
\label{sec:privacy_grad}
We test {\texthide} against the gradients matching attack in federated learning ~\cite{zlh19}, which has been shown effective in recovering private inputs from public gradients.

\paragraph{Gradients matching attack.} Given a public model and the gradients generated by private data from a client, the attacker aims to recover the private data: he starts with some randomly initialized dummy data and dummy labels (i.e., a dummy batch). In each iteration of attack, he calculates the $\ell_2$-distance between gradients generated by the dummy batch and the real gradients, and backpropagates that loss to update the dummy batch (see Algorithm~\ref{alg:attack_grad} in Appendix~\ref{sec:app_exp} for details).

The original attack is infeasible in the {\texthide} setting, because the attacker can't backpropagate the loss of the dummy batch through the {\em secret mask} of each client. Thus, we enhance the attack by allowing the attacker to learn the mask: at the beginning of the attack, he also generates some dummy masks and back-propagates the loss of gradient to update them. 

\paragraph{Setup and metric.} We use the code\footnote{\href{https://github.com/mit-han-lab/dlg}{https://github.com/mit-han-lab/dlg}} of the original paper~\cite{zlh19} for evaluation. Due to the unavailability of their code for attacks in text data, we adapted their setting for computer vision (see Appendix~\ref{sec:app_exp} for more details). We use the success rate as the metric: an attack is said to be successful if the mean squared error between the original input and the samples recovered from gradients is $\le 0.001$.  We vary two key variables in the evaluation: $k$ and $d$, where $d$ is the dimensionality of the representation (768 for BERT$_{\rm base}$).

\paragraph{Test the leakage upper bound.}  We run the attack in a much easier setting for the attacker to test the upper bound of privacy leakage:
\begin{itemize}
\vspace{-2mm}
    \item The {\texthide} scheme uses a single mask throughout training (i.e., $m=1$).
    \vspace{-3mm}
    \item The batch size is 1.\footnote{The original paper~\cite{zlh19} pointed out that attacking a larger batch is more difficult.}
    \vspace{-3mm}
    \item The attacker knows the true label for each private input.\footnote{As suggested by~\newcite{zhao2020idlg}, guessing the correct label is crucial for success in the attack.}
\end{itemize}

\begin{figure}[t]
    \centering
    \subfloat[Success]{\includegraphics[width=0.235\textwidth]{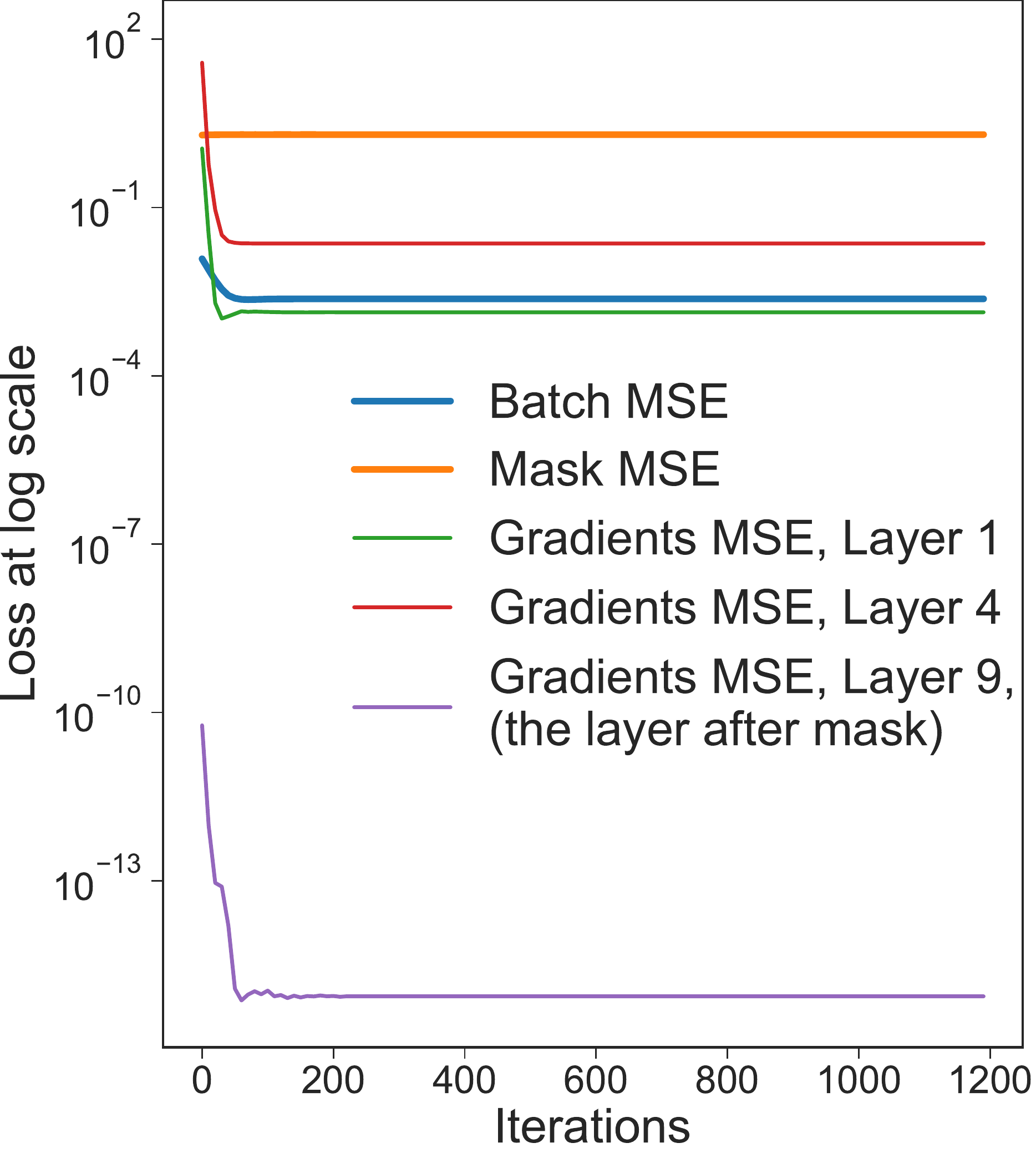}}
    \hspace{0.5mm}
    \subfloat[Failure]{\includegraphics[width=0.235\textwidth]{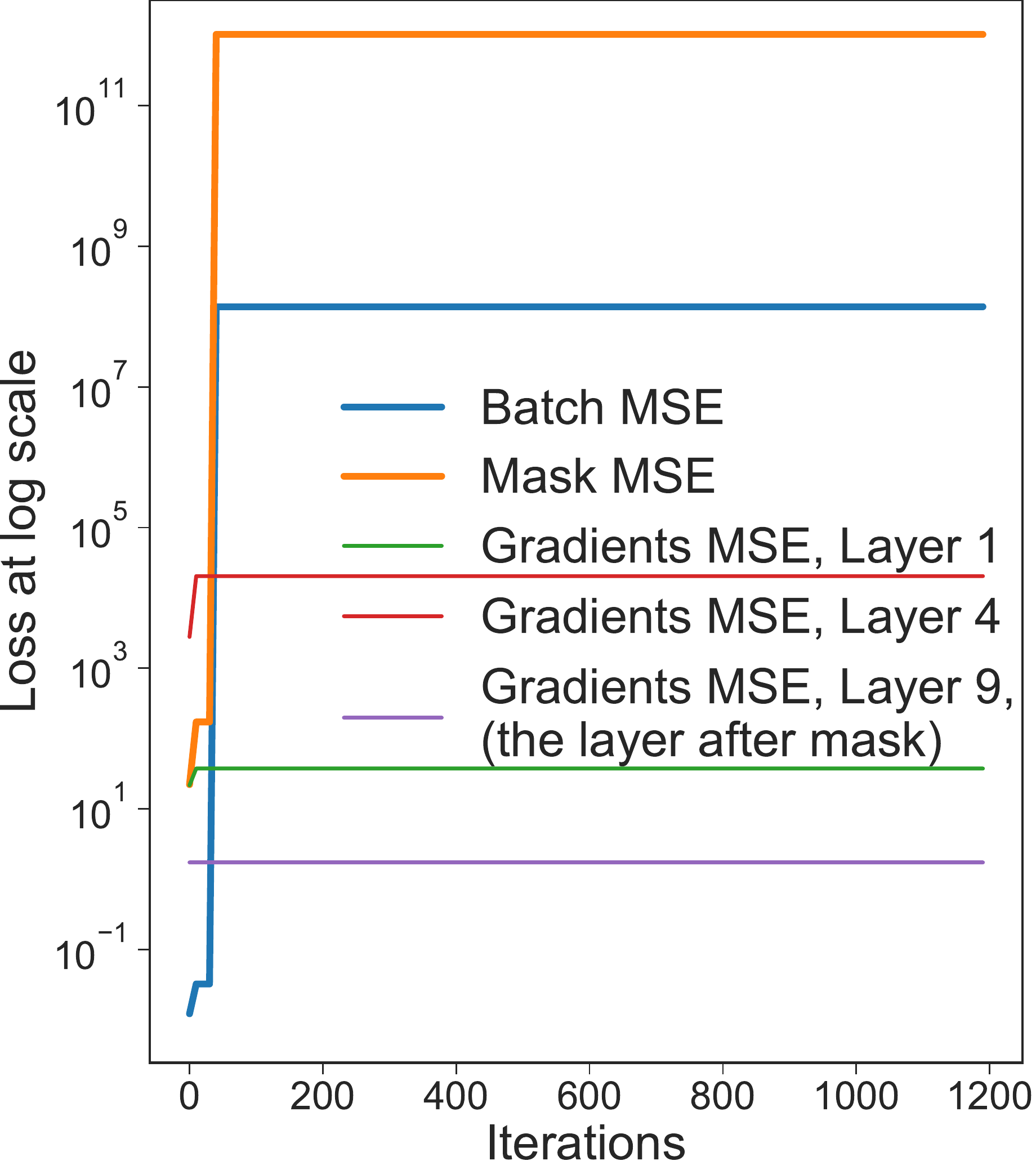}}
    \caption{Loss over iterations of a succeeded (a) and a failed (b) attacks. When the mean square error (MSE) between real and dummy masks gets smaller, both the gradients' distance and the MSE between leaked image and the original image gets smaller. 
    }
    \label{fig:dlg}
    \vspace{-2mm}
\end{figure}

\paragraph{{\texthide} makes gradients matching harder.} As shown in Table~\ref{tab:exp_dlg}, increasing $d$, greatly increases the difficulty of attack --- for no mixing ($k=1$), a representation with $d=1024$ reduces the success rate of $82\%$ (baseline) to only $8\%$. The defense becomes much stronger when combined with mixing: a small mask of 4 entries combined with $k=2$ makes the attack infeasible in the tested setting.
Figure~\ref{fig:dlg} suggests that the success of this attack largely depends on whether the mask is successfully matched, which is aligned with the security argument of {\texthide} in Section~\ref{sec:app_sec}. 

\begin{table}[!t]
    \small
    \centering 
    \begin{tabular}{c|c|cccccc}
    \toprule
    {\bf Baseline} & \diagbox[innerwidth=5.5mm]{$k$}{$d$}  & $4$& $16$ & $64$ & $256$ & $ 1024$\\
    \midrule
            & $1$&  0.76 & 0.56 & 0.30 & 0.22 & 0.08\\
      0.82  & $2$ &0.00 & 0.00 & 0.00 & 0.00& 0.00 & \\
            & $4$ & 0.00  & 0.00 & 0.00 & 0.00& 0.00 \\
    \bottomrule
    \end{tabular}
    \caption{Success rate of 50 independent gradients matching attacks. Baseline is the vanilla architecture without {\texthide}.
    $d$: the dimensionality of the representation. Increasing $k$ and $d$ makes attack harder.}
    \label{tab:exp_dlg}
    \vspace{-3mm}
\end{table}

\subsection{Effectiveness of Hiding Representations}
\label{sec:privacy_embed}
We also design an attack-based evaluation to test whether {\texthide} representations effectively ``hide'' its original representations, i.e., how `different' the {\texthide} representation is from its original representation. In Appendix~\ref{sec:app_exp}, we present another attack, which suggests that a deep architecture can not be trained to reconstruct the original representations from the {\texthide} representation.

\paragraph{Representation-based Similarity Search (RSS).}
Given a corpus of size $n$, and

\begin{table}[t]
\centering
\setlength{\tabcolsep}{5.5pt}
\subfloat[CoLA]{
\begin{tabular}{l|ccc|c}
\toprule
    & {\small Baseline} & {\em \small Mix-only} & {\em \small TextHide} & {\small Rand}\\
\midrule
    {Identity} & 0.993 & 0.111 & {\bf 0.000} & 0.000\\
    {JC$_{\rm dist}$} & 0.999 & 0.184 & {\bf 0.023} & 0.024 \\
    {TF-IDF$_{\rm sim}$} & 1.000 & 0.194 & {\bf 0.015} &  0.015\\
    {Label} & 0.998 & 0.759 & {\bf 0.494} & 0.542\\
    {SBERT$_{\rm sim}$} & 0.991 & 0.280 & {\bf 0.102} & 0.101\\
\bottomrule
\end{tabular}}\\
\vspace{-2mm}
\subfloat[SST-2]{
\begin{tabular}{l|ccc|c}
\toprule
    & {\small Baseline} & {\em \small Mix-only} & {\em \small TextHide} & {\small Rand}\\
\midrule
    {Identity} & 0.992 & 0.064 & {\bf 0.000} & 0.000\\
    {JC$_{\rm dist}$} & 0.999 & 0.168 & {\bf 0.100} & 0.096\\
    {TF-IDF$_{\rm sim}$} & 1.000 & 0.080&  {\bf  0.007} & 0.008 \\
    {Label} & 1.000 & 0.714 & {\bf 0.503} & 0.501\\
    {SBERT$_{\rm sim}$} & 1.000 & 0.275 & {\bf 0.202} & 0.209\\
\bottomrule
\end{tabular}}\\
\caption{Averaged similarity score of five metrics over 1,000 independent RSS attacks on CoLA (a) and SST-2 (b). For each score, the scheme with the worst similarity (best hiding) is marked in {\bf bold}. Rand: random baseline. As shown, attacker against {\texthide} gives similar performance to random guessing. 
}
\vspace{-2mm}
\label{tab:sim_metric}
\end{table}

\begin{table}[t]
\centering
\small
{\begin{tabular}{p{7cm}}
    \toprule
    {\bf Query1 (CoLA): \bf {\bf \color{mygreen} Some} {\bf \color{teal} people} {\bf \color{violet} consider} {\bf \color{magenta} the} {\bf \color{orange} noisy} {\bf \color{red}dogs} {\bf \color{pink} dangerous}.} (\checkmark)\\
    \midrule
    {\em Baseline}: ~\bf {\bf \color{mygreen} Some} {\bf \color{teal} people} {\bf \color{violet} consider} {\bf \color{magenta} the} {\bf \color{orange} noisy} {\bf \color{red}dogs} {\bf \color{pink} dangerous}. (\checkmark)\\
    {\em Mix-only}: ~~\bf {\bf \color{mygreen} Some} {\bf \color{teal} people} {\bf \color{violet} consider} {\bf \color{magenta} the} {\bf \color{orange} noisy} {\bf \color{red}dogs} {\bf \color{pink} dangerous}. (\checkmark)\\
    {\em TextHide}: I know a man who hates myself{\bf .} ($\times$)\\
    \midrule
    {\bf Query2 (SST-2): {\bf \color{mygreen} otherwise} {\bf \color{teal}{excellent}} } (\Laughey)\\
    \midrule
    {\em Baseline}: ~{\bf \color{mygreen} otherwise} {\bf \color{teal}{excellent}} (\Laughey)\\
    {\em Mix-only}: ~{\bf \color{teal} worthy} (\Laughey)\\
    {\em TextHide}: passive-aggressive (\Sadey)\\
    \bottomrule
    \end{tabular}}\\
    \caption{Example queries and answers of RSS with different representation schemes. We mark words with similar meanings in the same color. We annotate the acceptability for CoLA (`\checkmark': yes, `$\times$': no) and sentiment for SST-2 (`\Laughey': positive, `\Sadey': negative). Querying with a {\em Mix-only} representation still retrieve the original sentence (Query1), or sentence with similar meanings (Query2).}
    \vspace{-2mm}
\label{tab:sim_eval}
\end{table}

\begin{enumerate}
\vspace{-2mm}
    \item[1)] a search index: $\{x_i, e_i\}_{i=1}^n$, where $x_i$ is the $i$-th example in the training corpus
    , $e_i$ is $x_i$'s encoded representation $f_{\theta_1}(x_i)$;
    \vspace{-2mm}
    \item[2)] a query $\wt{e}$: {\texthide} representation of any input $x$ in the corpus,
\end{enumerate}

RSS returns $x_v$ from the index such that $v = \arg\min_{i\in[n]}{\cos}(e_i, \wt{e})$.
If $x_v$ is dramatically different from $x$, then $\tilde{e}$ hides $e$ (the original representation of $x$ ) effectively. To build the search index, we dump all $(x_i, e_i)$ pairs of a corpus by extracting each sentence's {\cls} token from the baseline BERT model. We use Facebook's FAISS library \cite{JDH17} for efficient similarity search to implement RSS.

\vspace{-2mm}
\paragraph{Metrics.} The evaluation requires measuring the similarity of a sentence pair, ($x,  x^*$), where $x$ is a sample in corpus, and $x^*$ is RSS's answer given $x$'s encoding $\tilde{e}$ as query. Our evaluation uses three explicit leakage metrics:
\begin{itemize}
\vspace{-2mm}
    \item {Identity}: 1 if $x^*$ is identical to $x$, else 0. 
    \vspace{-3mm}
    \item {JC$_{\rm dist}$}: Jaccard distance $|{\rm words~in~} x \cap {\rm words~in~} x^*| / |{\rm words~in~} x \cup {\rm words~in~} x^*|$
    \vspace{-3mm}
    \item {TF-IDF$_{\rm sim}$}: cosine similarity between $x$'s and $x^*$'s TF-IDF representation in the corpus
    \vspace{-5mm}
\end{itemize}

and two implicit (semantic) leakage metrics:
\begin{itemize}
\vspace{-2mm}
    \item {Label}: 1 if $x^*$, $x$ have the same label, else 0.
    \vspace{-3mm}
    \item {SBERT$_{\rm sim}$}:  cosine similarity between $x$'s and $x^*$'s SBERT representations pretrained on NLI-STS\footnote{We use SBERT as an off-the-shelf similarity scorer since it has been demonstrated great performance in semantic textual similarity tasks.} \cite{reimers2019sentence}.
\end{itemize}

For all five metrics above, a larger value indicates a higher similarity between $x$ and $x^*$, i.e., worse `hiding'.

\paragraph{Test Setup.} For an easier demonstration, we run RSS on two single-sentence datasets CoLA and SST-2 with {\hideintra}.
The results presumably can generalize to larger datasets and {\hideinter}, since attacking a small corpus with a weaker security is often easier than attacking a larger one with a stronger security. For each task, we test three $(m, k)$ variants:  baseline ($m=0, k=1$), mix-only ($m=0, k=4$), and {\texthide} ($m=256, k=4$). We report a random baseline for reference --- for each query, the attacker returns an input randomly selected from the index.

\paragraph{Baseline.} The result with original representation as query can be viewed as an upper bound of privacy leakage where no defense has been taken. As shown in Table~\ref{tab:sim_metric} and Table~\ref{tab:sim_eval}, RSS almost returns the correct answer all the time (i.e., \textit{Identity} close to 1), which is a severe explicit leakage.

\paragraph{Mix-only.} {\em Mix-only} representation greatly reduces both explicit leakage (i.e., gives much lower similarity on all first 3 metrics) compared to the undefended baseline. However, RSS still can query back the original sentence with Mix-only representations (see Query1 in Table~\ref{tab:sim_eval}). Also, semantic leakage, measured by \textit{Label} and \textit{SBERT$_{\rm sim}$}, is higher than the random baseline.

\paragraph{{\texthide}} {\texthide} works well in protecting both explicit and semantic information: sample attacks on {\texthide} (see Table~\ref{tab:sim_eval}) return sentences seemingly irrelevant to the original sentence hidden in the query representation. Note that the sophisticated attacker (RSS) against {\texthide} gives similar performance to a naive random guessing attacker.
\section{Related Work}

\vspace{-2mm}
\paragraph{Differential privacy.} 
Differential privacy \cite{dkmmn06,dr14} adds noise drawn from certain distributions to provide guarantees of privacy. Applying differential privacy techniques in distributed deep learning is interesting but non-trivial. \newcite{ss15} proposed a distributed learning scheme by directly adding noise to the shared gradients. \newcite{abadi2016deep} proposed to dynamically keep track of privacy spending based on the composition theorem~\cite{d09}, and ~\newcite{mcmahan2018learning} adapted this approach to train large recurrent language models. However, the amount of privacy guaranteed drops with the number of training epochs and the size of shared parameters \cite{papernot2020making}, and it remains unclear how much privacy can still be guaranteed in practical settings.

\vspace{-2mm}
\paragraph{Cryptographic methods.}  
Homomorphic encryption~\cite{g09,graepel2012ml,li2017multi} or secure multi-party computation (MPC)~\cite{yao82,b11,mpzy17,dglms19} allow multiple data cites (clients) to jointly train a model over their private inputs in distributed learning setting.
Recent work proposed to use cryptographic methods to secure federated learning by designing a secure gradients aggregation protocol \cite{bonawitz2017practical} or encrypting gradients \cite{ahwm17}. However, these approaches shared the same key drawback: slowing down the computation by several orders of magnitude, thus currently impractical for deep learning. 

\paragraph{{\instahide}.} See Section~\ref{sec:motivation}.

\paragraph{Privacy in NLP.} Training with user-generated language data raises privacy concerns: sensitive information can take the form of key phrases explicitly contained in the text ~\cite{harman2012electronic, hard2018federated}; it can also be implicit~\cite{cnc18, pan2020privacy}, e.g., text data contains latent information about the author and situation~\cite{hovy2016social, elazar2018adversarial}. Recently, ~\newcite{song2020information} suggests that text embeddings from language models such as BERT can be inverted to partially recover some of the input data.

To deal with explicit privacy leakage in NLP,
\newcite{zhang2018privacy} added DP noise to TF-IDF~\cite{salton1986introduction} textual vectors, and~\newcite{hhtc19} obfuscated the text by substituting each word with a new word of similar syntactic role. However, both approaches suffer large utility loss when trying to ensure practical privacy. 

Adversarial learning~\cite{lbc18, hhtc19} has been used to address implicit leakage to learn representations that are invariant to private-sensitive attributes. Similarly, ~\newcite{mbl19} used reinforcement learning to automatically learn a strategy to reduce private-attribute leakage by playing against an attribute-inference attacker. However, these approaches does not defend explicit leakage.
\section{Conclusion}
\vspace{-2mm}
We have presented {\placeholder}, a practical approach for privacy-preserving NLP training with a pretrain and fine-tuning framework in a federated learning setting. It requires all participants to add a simple encryption step with an one-time secret key. It imposes a slight burden in terms of computation cost and accuracy. Attackers who wish to break such encryption and recover user inputs have to pay a large computational cost. 

We see this as a first step in  using cryptographic ideas to address privacy issues in language tasks.  We  hope our work motivates further research, including applications to other NLP tasks. An important step could be to successfully train language models directly on encrypted texts, as is done for image classifiers.


\section*{Acknowledgements}

This project is supported in part by the Graduate Fellowship at Princeton University, Ma Huateng Foundation, Schmidt Foundation, Simons Foundation, NSF, DARPA/SRC, Google and Amazon AWS. Arora and Song were at the Institute for Advanced Study during this research.
\ifdefined\isarxivversion
\bibliographystyle{alpha}
\else
\bibliographystyle{acl_natbib}
\fi
\bibliography{ref}

\clearpage
\appendix

\section{$k$-{\sc vector subset sum}}
\label{sec:app_ksum}
Cryptosystem design since the 1970s seeks to ensure that attackers can violate privacy only by solving a computationally expensive task. A simple example  is the {\sc vector subset sum} problem \cite{biwx11,alw14}. Here a set of $N$ vectors $v_1, v_2, \ldots, v_k \in {\mathbb R}^d$ are publicly released. The defender picks secret indices $i_1, i_2,\ldots, i_k \in [N] \overset{\mathrm{def}}{=} \{1,\cdots,N\}$ and publicly releases the vector $\sum_{j} v_{i_j}$. Given this released vector the attacker has to find secret indices $i_1, i_2, \ldots, i_k$. In worst cases even when the answer happens to be unique, finding the secret indices requires $\geq N^{k/2}$ time \cite{al13} under the famous conjecture, Exponential Time Hypothesis (\ETH) \cite{ipz98}. Note that {\ETH} is a stronger notion than $\mathsf{NP} \neq \mathsf{P}$, and {\ETH} is widely accepted computational complexity community.
\section{Experiment details}

\subsection{Implementation}

The implementation uses the PyTorch framework \cite{pytorch19} based on HuggingFace’s codebase \cite{wolf2019transformers}. We ran all experiments on 24 NVIDIA RTX 2080 Ti GPUs.

\subsection{More Evaluations}
\label{sec:app_more_exp}

\begin{table*}[t]
\setlength{\tabcolsep}{1.5pt}
    \centering
    \begin{tabular}{lccccccccc}
    \toprule
    {\bf Datasets} & $|\cal D|$ & {\bf Task} & {\bf Metric} & {\bf Baseline} & {\bf \hideintra} & {\bf \hideinter}\\
    \midrule
    {\bf RTE} &  2.5k & NLI & Acc. & $78.8_{(0.69)}$ & $77.9_{(0.99)}$ & $70.8_{(0.78)}$ \\
    {\bf MRPC} & 3.7k & Paraphrase &  F1 / Acc. & $92.3_{(0.56)}$ / $89.3_{(0.66)}$ & $91.1_{(0.68)}$ / $87.6_{(0.34)}$ & $90.5_{(0.68)}$ / $87.1_{(0.89)}$\\
    {\bf STS-B} &  7k & Similarity & P / S corr. & $91.3_{(0.13)}$ / $91.0_{(0.19)}$ & $90.4_{(0.19)}$ / $90.3_{(0.16)}$ & $82.6_{(0.65)}$ / $84.2_{(0.52)}$\\
    {\bf CoLA} &  8.5k & Acceptability & MCC & $63.0_{(1.24)}$ & $59.1_{(1.01)}$ & $57.2_{(0.90)}$\\
    {\bf SST-2} & 67k & Sentiment & Acc. & $94.1_{(0.52)}$ & $93.5_{(0.21)}$ & $92.8_{(0.47)}$\\
    {\bf QNLI} & 108k & NLI & Acc. & $92.7_{(0.21)}$ & $92.3_{(0.29)}$ & $91.7_{(0.48)}$\\
    {\bf QQP} & 364k & Paraphrase & F1 / Acc. & $88.8_{(0.21)}$ / $91.6_{(0.15)}$ & $88.1_{(0.24)}$ / $91.0_{(0.31)}$ & $87.7_{(0.36)}$ / $90.7_{(0.22)}$\\
    {\bf MNLI} & 393k & NLI & m/mm & $87.2_{(0.39)}$ / $86.8_{(0.21)}$ & $86.4_{(0.21)}$ / $86.0_{(0.15)}$ & -\\
    \bottomrule
    \end{tabular}
    \caption{Performance on the GLUE tasks for both baseline (standard finetuning) and {\placeholder} with RoBERTa$_{\rm base}$ model~\cite{liu2019roberta}, measured on the development sets. We report the mean results across 5 runs, with  $(m,k)=(16,4)$ for RTE and $(m,k)=(256,4)$ for all the other datasets. Standard deviations are reported in parentheses. $|\cal D|$ denotes the number of training samples. {\texthide} only suffers minor utility loss ($\sim3\%$). 
    `P / S corr.' is Pearson/Spearman correlation. `MCC' is Matthew's correlation.
    }
    \label{tab:summary_roberta}
\end{table*}

\begin{table}[t]
\small
    \centering
    \begin{tabular}{c|cccc}
    \toprule
    \multicolumn{4}{c}{{\bf Private dataset: SST-2} ($|\cal D|$: 67k)} \\
    \midrule
         {\bf Public Corpora} & $|\cal D|$ & {\bf Task}& {\bf Acc.} \\
    \midrule
        QNLI & 108k & NLI & $91.2_{(0.68)}$\\
        QQP & 364k  & Paraphrase & $91.0_{(0.45)}$\\
        MNLI & 393k & NLI & $91.3_{(0.41)}$ \\
    \bottomrule
    \end{tabular}
    \caption{Dev set performance of SST-2 for {\hideinter} with different public corpora, $(m, k)=(256, 4)$. $|\cal D|$ denotes the number of samples. Standard deviations are annotated as subscripts. The choice of the public corpus does not have a major impact on the final accuracy of SST-2.}
    \label{tab:pub_test}
\end{table}

\paragraph{Compatibility with the state-of-the-art model.} To test if {\texthide} is also compatible with state-of-the-art models, we repeat our accuracy evaluation in Section~\ref{sec:utility} but replace the BERT$_{\rm base}$ model with the RoBERTa$_{\rm base}$ model~\cite{liu2019roberta}. 

As shown in Table~\ref{tab:summary_roberta}, {\texthide} behaves consistently for BERT$_{\rm base}$ and RoBERTa$_{\rm base}$: when incorporated with RoBERTa$_{\rm base}$, the averaged accuracy reduction of {\hideintra} is $1.1\%$ when compared with the baseline model (was $1.9\%$ for BERT$_{\rm base}$). {\hideinter} incurs an additional $2.6\%$ accuracy loss on average (was $2.5\%$ for BERT$_{\rm base}$).

\paragraph{{\hideinter} with different public corpora: A case study of SST-2.} 

We investigate whether using different public corpora affects the performance of {\hideinter}. We fix SST-2 as the private dataset, set $(m, k)=(256,4)$, and choose the public corpora from {\em unlabeled} $\text{\{QNLI, QQP, MNLI\}}$. We intentionally make the public corpora larger than the private dataset (SST-2 in this test), since {\hideinter} was designed to use a {\em large} public corpus as the source of randomness to provide useful security.

Table~\ref{tab:pub_test} suggests that for our case study of SST-2, the choice of the public corpus does {\em not} have a major impact on the final accuracy of {\hideinter}. However, this may not be true for every dataset.

\subsection{Fine-tuning Hyperparameters}
\label{sec:app_hyperparam}

For results in Table~\ref{tab:summary_exp} and \ref{tab:summary_roberta} (including our baseline), we chose the best parameters with learning rate = $\{5e-6, 1e-5, 2e-5, 3e-5, 5e-5\}$, epochs = $\{5, 10, 15, 20, 25, 30\}$, batch size = $\{16, 32\}$, dropout rate = $\{0.1, 0.2, 0.3, 0.4, 0.5\}$ based on the validation performance ($10\%$ from the training set). We used more epochs for fine-tuning since training with random masking takes longer to converge.

\section{Details of attacks}
\label{sec:app_exp}

\begin{algorithm}[t]
\small
\caption{Gradients matching attack~\cite{zlh19} in {\texthide}}
  \begin{algorithmic}[1]
  \State {\bf Require : }
  \State The function $F(x; W)$ can be thought of as a neural network 
  \State For each $l \in [ L ]$, we define $W_l \in \R^{m_l \times m_{l-1}}$ to be the weight matrix in $l$-th layer, and $m_0 = d_i$ and $m_l = d_o$
  \State Let $W = \{W_1, W_2, \cdots, W_L\}$ denote the weights over all layers 
  \State Let $\mathcal{L} : \R^{d_o \times d_o} \rightarrow \R$ denote loss function
   \State Let $g(x,y) = \nabla {\cal L} ( F(x;W) ,y)$ denote the gradients of loss function
  \State Let $\hat{g} =  g(\sigma, x,y) |_{ \sigma= \sigma_0, x =x_0, y= y_0 }$ denote the gradients computed on $x_0$ with label $y_0$, and secret mask $\sigma_0$
  \label{line:real_grad}
  \Procedure{\textsc{InputRecoveryfromGradients}}{}
      \State $x^{(1)} \gets \mathcal{N} (0, 1)$, $y^{(1)} \gets \mathcal{N} (0, 1)$, $\sigma^{(1)} \gets \mathcal{N} (0, 1)$
      \Comment{Random initialization of the input, label and mask}
      \label{line:dummy_mask}
      \For{$t = 1 \rightarrow T$}
     \State Let $D_g(\sigma, x,y) = \|g(\sigma, x,y) - \hat{g} \|_2^2$
      \State $x^{(t+1)} \gets x^{(t)} - \eta \cdot \nabla_{x} D_g(\sigma, x,y) |_{x = x^{(t)}} $
      \State $y^{(t+1)} \gets y^{(t)} - \eta \cdot \nabla_{y} D_g(\sigma, x,y) |_{ y=y^{(t)} } $
      \State $\sigma^{(t+1)} \gets \sigma^{(t)} - \eta \cdot \nabla_{y} D_g(\sigma, x,y) |_{\sigma=\sigma^{(t)} } $
      \label{line:update_mask}
      \EndFor
  \State \Return $x^{(T+1)}$, $y^{(T+1)}$, $\sigma^{(T+1)}$
  \EndProcedure
  \end{algorithmic}
  \label{alg:attack_grad}
  \end{algorithm}

\begin{table}[!t]
\centering
\small
\begin{tabular}{p{7cm}}
\Xhline{0.8pt}
    {\bf Q1(CoLA): {\bf \color{mygreen}The} {\bf \color{teal} magazines} {\bf \color{b2} were} {\bf \color{violet} sent} {\bf \color{magenta} to} {\bf \color{orange} herself} {\bf \color{red}by} {\bf \color{pink} Mary}}{\bf.} ($\times$)\\
    \midrule
    {\sf Baseline}: ~{\bf \color{mygreen}The} {\bf \color{teal} magazines} {\bf \color{b2} were} {\bf \color{violet} sent} {\bf \color{magenta} to} {\bf \color{orange} herself} {\bf \color{red}by} {\bf \color{pink} Mary}{\bf.} ($\times$)\\
    {\sf Mix-only}: ~{\bf \color{mygreen}The} company {\bf \color{violet} sent} China its senior mining engineers {\bf \color{magenta} to} help plan {\bf \color{mygreen}the} new mines. ($\times$)\\
    {\sf TextHide}: Hierarchy of Projections: (\checkmark)\\
    \midrule
    {\bf Q2(SST-2): {\bf \color{mygreen} an}  {\bf \color{teal} exquisitely} {\bf \color{b2} crafted}  {\bf \color{violet} and}  {\bf \color{magenta} acted} {\bf \color{orange} tale}{\bf.}} (\Laughey) \\
    \midrule
    {\sf Baseline}: {\bf \color{mygreen} an}  {\bf \color{teal} exquisitely} {\bf \color{b2} crafted}  {\bf \color{violet} and}  {\bf \color{magenta} acted} {\bf \color{orange} tale}{\bf.} (\Laughey)\\
    {\sf Mix-only}:~ to make their way through this tragedy (\Sadey)\\
    {\sf TextHide}: fails to live up to -- or offer any new insight into -- its chosen topic (\Sadey)\\
    \bottomrule
    \end{tabular}
    \caption{Example queries and answer of RepRecon with different representation schemes. Words with similar meanings are marked in the same color. For CoLA examples, we annotate the acceptability (`\checkmark' for yes, `$\times$' for no); for SST-2 examples, we annotate sentiment (`\Laughey' for positive, `\Sadey' for negative).}
    \label{tab:sim_eval_app}
\end{table}

\begin{table}[!t]
\small
\centering
\subfloat[RepRecon, CoLA]{
\begin{tabular}{l|ccc||c}
\Xhline{0.8pt}
    & {\sf \scriptsize Baseline} & {\sf \scriptsize Mix-only} & {\sf \scriptsize TextHide} & {\sf \scriptsize Rand}\\
\Xhline{0.8pt}
    {\bf ID} & 0.982 & 0.002 & {\bf 0.000} &  0.000\\
    {\bf JC$_{\rm dist}$} & 0.992 & 0.033 & {\bf 0.029} & 0.028\\
    {\bf TF-IDF$_{\rm sim}$} & 0.993 & 0.018 & {\bf 0.014} & 0.018 \\
    {\bf Label} & 0.998 & 0.818 & {\bf 0.638} & 0.620\\
    {\bf SBERT$_{\rm sim}$} & 0.994 & 0.111 & {\bf 0.051} & 0.104\\
\Xhline{0.8pt}
\end{tabular}}\\

\subfloat[RepRecon, SST-2]{
\begin{tabular}{l|ccc||c}
\Xhline{0.8pt}
    & {\sf \scriptsize Baseline} & {\sf \scriptsize Mix-only} & {\sf \scriptsize TextHide} & {\sf \scriptsize Rand}\\
\Xhline{0.8pt}
    {\bf ID} & 0.948 & {\bf 0.000} &  {\bf 0.000} &  0.000\\
    {\bf JC$_{\rm dist}$} & 0.953 & 0.065 & {\bf 0.064} & 0.080\\
    {\bf TF-IDF$_{\rm sim}$} & 0.949 & 0.019 & {\bf 0.013} & 0.014 \\
    {\bf Label} & 0.968 & 0.464 & {\bf 0.472} & 0.452\\
    {\bf SBERT$_{\rm sim}$} & 0.959 & 0.268 & {\bf 0.266} &  0.211\\
\Xhline{0.8pt}
\end{tabular}}
\caption{Similarity score of five metrics for RepRecon on CoLA (a) and SST-2 (b) datasets. We report the average score over 500 independent queries. Test queries come from only the dev set. For each score, the scheme with the worst similarity (best hiding) is marked in {\bf bold}. As shown, attacker against {\texthide} gives similar performance to random guessing.}
\label{tab:sim_metric_app}
\end{table}

\subsection{Gradients matching attack}
Algorithm \ref{alg:attack_grad} describes the gradients matching attack \cite{zlh19} in TextHide setting. This attack aims to recover the original image from model gradients computed on it. As discussed in Section~\ref{sec:motivation}, masks are kept private in TextHide setting, thus the attacker also need to start from a dummy mask (line~\ref{line:dummy_mask}) and iteratively update it to compromise the real mask (line~\ref{line:update_mask}). In our experiment, we made this attack much easier for the attacker, by revealing to him the real ground truth label ($y_0$ in line~\ref{line:real_grad}), which means he simply sets $y^{(t)} = y_0$ throughout the attack. 

\paragraph{Dataset and architecture.} We used CIFAR-10~\cite{cifar10} as the dataset and LeNet-5~\cite{lecun1998gradient} as the architecture to mimic {\texthide}.

Given the original LeNet-5, we firstly removed the last linear layer with output size $d_o$, which gives us a new network. We use $d_c$ to denote the size of output in the new network. Then, we appended an MLP with hidden-layer size $d_m$ and output size $d_o$ to the new architecture. As in an $(m,k)$-{\texthide} scheme, for each private input, we first gets its {\texthide} representations by extracting the output from the hidden-layer, and mixes it with representations of other datapoints. We then apply a mask on this combination. Note: in this mimic setting, the mask's dimension is $d_m$.

\paragraph{Hyper-parameters and running-time.} Following~\cite{zlh19}, we use L-BFGS~\cite{liu1989limited} optimizer (learning-rate 1, history-size 100 and max-iterations 20) and optimize for 1,200 iterations. Each run takes 97 seconds (single V100 GPU, averaged across 20 runs).

\subsection{Representation-based Similarity Search (RSS)}
\paragraph{Running-time.} For CoLA, building the search index takes $267$ seconds; each search takes $<0.1$ seconds. For SST-2, building the index takes $1,576$ seconds; each search takes $<0.1$ seconds.

\subsection{Representation Reconstruction (RepRecon)}

RepRecon tests whether a deep architecture can learn to disrupt our `hiding' scheme. For an representation $e \in \R^d$, and its {\texthide} version $\wt{e} \in \R^d$, RepRecon tries to reconstruct $e$ from $\wt{e}$ by training a network $f : \R^d \rightarrow \R^d$ such that $ \|e - f(\wt{e}) \|_2$ is minimized. 

We use a multi-layer perception of hidden-layer size (1024, 1024) as the reconstruction architecture. We train the network on the train set of a benchmark for 20 epochs, and run evaluation using the dev set. We then run RSS to map the recovered representation to its closet sentence in the index, and measure the privacy leakage.

Quantitative and qualitative results of RepRecon are shown in Table~\ref{tab:sim_metric_app} and Table~\ref{tab:sim_eval_app}.

\end{document}